\documentclass[5p]{elsarticle}
\pdfoutput=1
\usepackage[cmex10]{amsmath}
\usepackage[font=footnotesize]{subfig}
\usepackage{algorithm,algpseudocode}
\usepackage{subfig}
\usepackage{graphicx}
\usepackage{amssymb}

\newcommand{\set}[1]{\mathbb{#1}}

\newtheorem{defi}{Definition}
\newtheorem{lem}{Lemma}

\newtheorem{thm}{Theorem}
\newtheorem{ass}{Assumption}
\newproof{pf}{Proof}
\newcommand{\defref}[1]{Definition~\ref{#1}}
\newcommand{\figref}[1]{Figure~\ref{#1}}

\newcommand{\secref}[1]{Section~\ref{#1}}
\newcommand{\algoref}[1]{Algorithm~\ref{#1}}

\newcommand{\lemref}[1]{Lemma~\ref{#1}}
\newcommand{\assref}[1]{Assumption~\ref{#1}}
\newcommand{\pijl}{\leftarrow}

\begin{document}

\begin{frontmatter}
%\title{Multi-robot coverage to locate fixed targets using formation structures}
%
%\author{, %~\IEEEmembership{Member,~IEEE,}
%        Dirk~Aeyels%~\IEEEmembership{Fellow,~OSA,}% <-this % stops a space
%\thanks{J. A. Rogge and D. Aeyels are with the SYSTeMS Research Group, Ghent University, Belgium
% e-mail: (see http://users.ugent.be/\~{}jarogge/address.html).}% <-this % stops a space
%%\thanks{Manuscript received April 19, 2005; revised January 11, 2007.}
%}

\title{Multi-robot coverage to locate fixed targets using formation structures}

\author[rvt]{J.A.~Rogge}
\ead{jonathan.rogge@ugent.be}
\author[rvt]{D.~Aeyels}
\ead{dirk.aeyels@ugent.be}
\address[rvt]{SYSTeMS Research Group, Ghent University, Technologiepark Zwijnaarde 914, 9052 Zwijnaarde, Belgium}

\begin{abstract}
This paper develops an algorithm that guides a multi-robot system in
an unknown environment in search of fixed targets. The area to be
scanned contains an unknown number of convex obstacles of unknown
size and shape. The algorithm covers the entire free space in a
sweeping fashion and as such relies on the use of robot formations.
The geometry of the robot group is a lateral line formation, which
is allowed to split and rejoin when passing obstacles. It is our
main goal to exploit this formation structure in order to reduce
robot resources to a minimum. Each robot has a limited and finite
amount of memory available. No information of the topography is
recorded. Communication between two robots is only possible up to a
maximum inter-robot distance, and if the line-of-sight between both
robots is not obstructed. Broadcasting capabilities and indirect
communication are not allowed. Supervisory control is prohibited.
The number of robots equipped with GPS is kept as small as possible.
Applications of the algorithm are mine field clearance,
search-and-rescue missions, and intercept missions. Simulations are
included and made available on the internet, demonstrating the
flexibility of the algorithm.
\end{abstract}

%\IEEEpeerreviewmaketitle

\end{frontmatter}
\section{Introduction}
\subsection{Multi-robot coverage based on single-robot algorithms}
The research domain of multi-agent mobile robot systems consists of
subdomains according to the task to be performed by the robot group
\cite{Ota}. At present, well-studied subdomains are motion-planning
(also called path-plan\-ning), formation-forming, region-sweeping,
and combinations thereof. This paper discusses region-sweeping.
Re\-gion-sweeping algorithms appear in two distinct types of robot
missions: mapping of an area and coverage of an area. A mapping
algorithm produces a topographical map of the area; a coverage
algorithm is performed successfully when all free space has been
covered by the robots, without demanding a map. We restrict our
attention to coverage problems where teams of robots are involved.
%
%The problem statement of the present paper belongs to the category
%of coverage missions using robot formations. Our aim is to locate
%all targets with the robots' sensors within an unknown area, with an
%absolute certainty that all free space has been covered by the
%sensors at the end of the procedure.

We briefly recall the most relevant approaches to multi-robot
coverage missions. The most basic approach lets the robots move
around independently in a random fashion \cite{Keymeulen,Spears}.
With time tending to infinity the entire area gets covered.

A more structured approach extends existent exploration algorithms
for a single robot to the multi-robot case with unlimited
communication capabilities. Unlimited communication can be realized
in two ways:
\begin{enumerate}
    \item every robot is able to broadcast messages to all other robots
    irrespective of inter-robot distance,
    \item every robot can exchange information with a shared memory
    unit.
\end{enumerate}
A typical strategy combining single-robot coverage techniques,
consists of dividing the exploration area into separate regions,
each of which is assigned to a single robot~\cite{Burgard}.
Efficient ways to assign these regions to the robots are discussed
in \cite{Agmon,Hazon,Zheng,Butler}. These algorithms use a
supervisor for the assignment and require a-priori knowledge on the
lay-out of the environment. Throughout the algorithm the robots
transmit information to each other (or a central shared memory unit)
with regard to the area they have covered so far, in order to
minimize the probability of covering the same area more than once,
reducing the operating time of the algorithm significantly.

Other research groups have developed robot systems that combine
single-robot coverage techniques as described above, but assume more
realistic situations with one or more of the following assumptions:
\begin{itemize}
    \item The communication and sensing properties of the robots are
limited,
    \item the environment is unknown,
    \item no supervisory control or shared memory is allowed.
\end{itemize}
In \cite{Luo} for instance, a strategy is used that does not require
a supervisor to distribute the area between robots: each robot is
guided by a neural network representing the (known) environment.
This network ensures the robot is globally attracted towards
unscanned areas. Cooperation among robots consists of collision
avoidance between them. In other words communication is limited to
robots with sufficiently small inter-robot distance.

A line of research considering robot coverage where all three of the
above assumptions hold, consists of so-called ant-robotics
\cite{Koenig,Wagner,Menezes}. Ant-robots use very limited
communication and need hardly any memory. Their behavior is inspired
by ant behavior as found in nature: the robots are able to leave
so-called pheromone traces in the environment. This pheromone serves
as a means of indirect communication replacing inter-robot
communication and a shared memory. The environment is divided into a
grid; pheromone is represented by a quantity that assumes a value in
every cell of the grid and determines the behavior of the robot in
the respective cell. In essence, the coverage strategy with
ant-robotics is identical to the strategy described before: each
robot in the team performs single-robot coverage. The pheromone
indicates which areas have already been covered, preventing the same
area to be covered twice. In experimental setups pheromone is
represented by a chemical substance, a heat trail, pen marks, or
crumbs. The choice needs to be adjusted to the complexity of the
algorithm under consideration. Implementation is not always
straightforward and is the subject of on-going research \cite{Kowadlo}.%%Referenties toevoegen! Russell,...

\subsection{Multi-robot coverage based on robot formations}
\label{sec:robform} There exists an approach to the coverage problem
which does not resort to transposing single-robot algorithms to a
multi-robot setting, but takes advantage of the cooperation
capabilities of a robot team. The robot team or its subgroups
maintain formations to scan the area
\cite{Kong,Latimer,Rekleitis2,Rekleitisjournal}. The degree of
rigidity of and cooperation inside the formations may vary depending
on the approach. The environment is represented by standard
Euclidean space instead of a static grid structure. A successful
covering approach with robot formations uses the so-called
boustrophedon decomposition \cite{Rekleitisjournal}. This is a
cellular decomposition of the unknown environment, where each cell
has the property that it can be scanned by back-and-forth motions of
the robot team. The robot team itself creates this decomposition in
an online fashion, i.e. during the algorithm's execution. Although
this approach is able to tackle an unknown environment without using
a supervisory control, a major draw-back remains the heavy use of
sensing, communication and memory capabilities of the robots. Every
robot needs to store the cellular decomposition of the environment
into its memory. It does this in the abstracted form of a so-called
Reeb graph. The topology of the Reeb graph depends on the shape and
position of the obstacles. Vertices of the graph correspond with
specific corners of the obstacles. The robots need to record the
coordinates of these points using GPS and store them into their
memory for later use. The online construction of a Reeb graph is not
straightforward, as explained in \cite{Garcia}. The robots exchange
information on the graph in order to obtain a global picture of the
environment, which enables them to keep track of areas left to be
covered. Moving to uncovered areas may happen over long distances
through already covered terrain.

%In some cases a robot group has to cross a cell which
%has already been treated by the algorithm to reach the next cell to
%be covered. In other words, these algorithms guarantee coverage of
%the entire area, but do not minimize the probability of covering the
%same area more than once.
%
%Reference~\cite{Choset} includes a short overview of other existing
%techniques for multi-robot coverage problems. Remaining approaches
%to the coverage problem are found in
%\cite{Anisi,Cortes,Kurabayashi,Solanas}.
\subsection{Our approach}
The present paper combines ideas from ant-robotics and coverage with
robot formations, leading to a novel approach of the coverage
problem. In line with the philosophy of ant-robotics where
simplicity of the robots is a key element (recall the three
assumptions on communication/sensing, the environment and
supervisory control mentioned above), this paper investigates the
possibility of significantly reducing the necessary resources of the
robots. Instead of resorting to indirect communication through
pheromone, we investigate the benefit brought about by robot
formations to guarantee coverage. The robots' resources are limited
in the following sense. The sensing capabilities of each robot are
of limited range. Inter-robot communication is allowed for robots
which are located sufficiently close to one another. Communication
with a shared memory unit or via pheromone is prohibited.
Decision-making is fully decentralized: each robot determines its
actions based on its own observations and memory content. The
lay-out of the environment to be covered is not known a-priori.
Moreover, to reduce the demands on each robot's memory, a map or
abstract representation of the explored environment is not stored in
any kind of memory device. The memory use of each robot is reduced
and does not scale with the size of the problem or the size of the
robot team. Finally, as few robots as possible know their absolute
position. To the best of our knowledge, no other approach in the
literature is able to reduce the robots' resources to the same
degree. However, putting all these limitations in place comes at a
cost: the algorithm constructed in the present paper guarantees full
coverage of spaces containing only convex obstacles. If we want to
be able to treat (subclasses of) nonconvex obstacles, we need to
relax the aforementioned limitations.

Our coverage method, a preliminary version of which has appeared in
\cite{RoggeMSC09}, does not apply a dynamic cellular decomposition
%which is determined by shape and position of encountered obstacles
as described in \secref{sec:robform}. We let the robot group sweep
the area in one or more predefined {\it scanning strips}. These
strips do not take the presence of obstacles into account: obstacles
are allowed to intersect the boundaries of adjacent strips. The
robot group sweeps the area strip by strip in a back-and-forth
fashion while dealing with the obstacles encountered inside the
strip. The size of the strips depends on the number of robots. The
location of the strips is communicated to the outer robots of the
formation at the onset of the algorithm. These two robots follow the
strip boundaries using a GPS system. The remaining robots stay in
formation which leads to a successful coverage of a strip. The
algorithm is described in detail in \secref{sec:alg}. We also
provide a proof that the algorithm guarantees full coverage (see
\secref{sec:proof}).
%the robot team are equipped with.

Some applications we have in mind are mine field clearance using
chemical vapor microsensors (introduced in  \cite{Gage}) and
search-and-rescue of snow avalanche victims, using specialized
transceivers. In both cases the robots are equip\-ped with
specialized sensors to locate the targets. When a target is detected
a signal is given to start negotiating the target. In case of mine
field clearance this would be dismantling the mine, in search and
rescue this is evacuating the victim.

\section{Defining the setting}

\subsection{Modeling the environment}\label{sec:modenv}
\begin{figure}[ht]
\centering
\includegraphics[width=8cm]{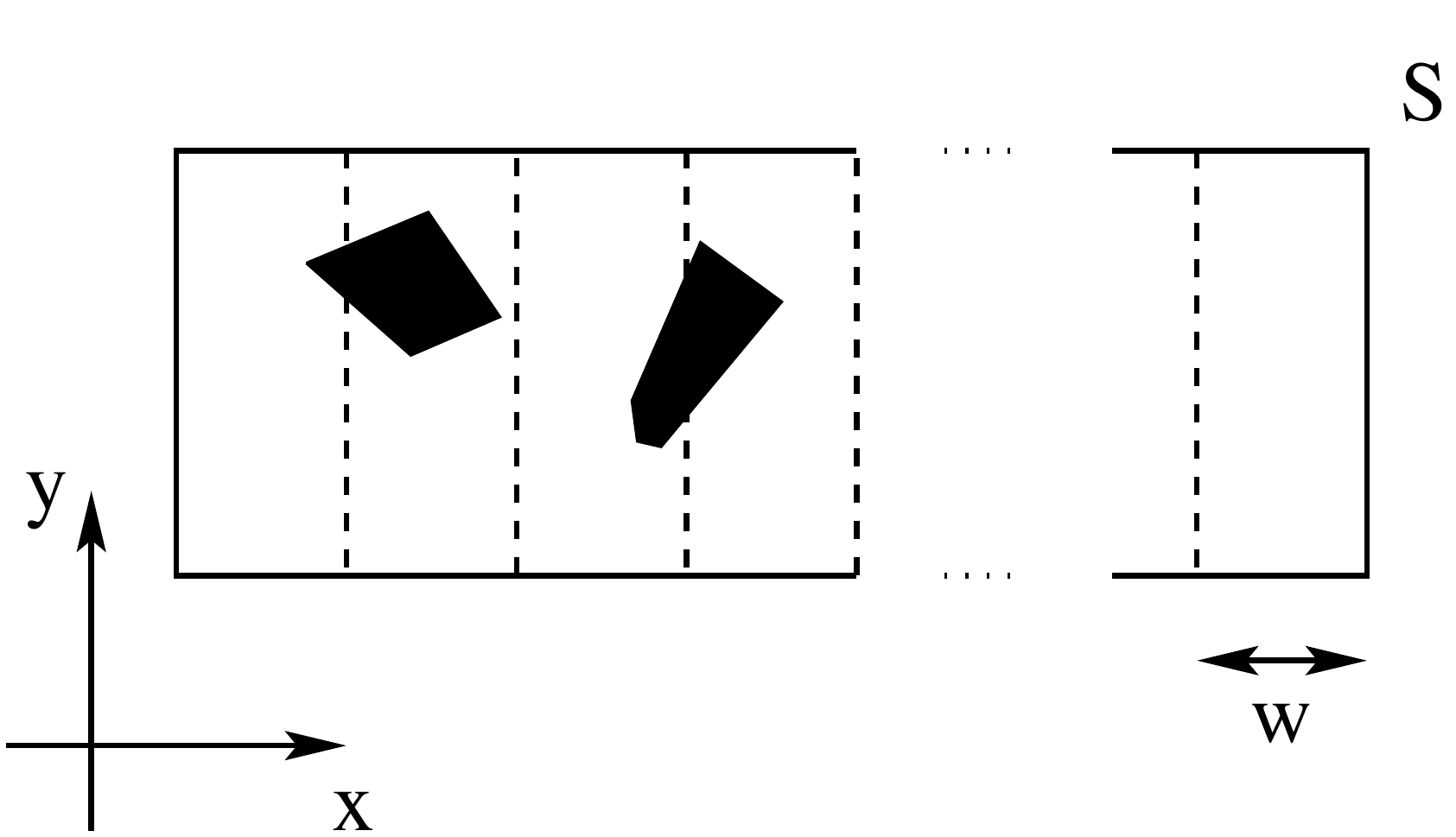}
\caption{The area $S$ assigned to the scanning algorithm; the black
polygons represent obstacles.} \label{fig:environment}
\end{figure}

We single out a rectangular area $S \subset \set{R}^2$ that we want
to explore. The set $S$ is divided into several parallel rectangles
with equal width $w$. These rectangles are called scanning strips.
We define a right-handed $(x,y)$-frame, with the $y$-axis directed
along the common boundaries of the scanning strips (see
\figref{fig:environment}).

Obstacles inside $S$ are represented by polygons. %A convex
%polygon $P$ is defined by a given set of points
%$p_i:=(p_{i_x},p_{i_y}) \in S \subset \set{R}^2, \: i=1,\ldots,m$,
%as follows:
%\begin{equation}
%\label{eq:obst} P := \{\sum_{k=1}^{m} \alpha_k p_k | \alpha_k \in
%\set{R}_{\geq 0}, \sum_{k=1}^{m} \alpha_k =1\}.
%\end{equation}
With $N_O$ denoting the number of obstacles in $S$, each obstacle
obtains an index $i_O \in \mathcal{N}_O := \{1,\ldots,N_O\}$.

The maximal diameter $\gamma$ of the obstacles is defined by
\begin{equation}
\gamma := \max_{i_O} \max_{p,q \in P_{i_O}} \Vert p-q \Vert,
\label{eq:maxdiameter}
\end{equation}
% (The diameter of an obstacle is defined as the greatest
%distance between any two points on the boundary of the obstacle.)
with $P_{i_O}$ an obstacle in $S$. Furthermore it is assumed that
the distance between obstacles is sufficiently large. Further on in
the paper a lower bound on the inter-obstacle distance is given (see
\secref{sec:siteentwee}).

%Except for \secref{sec:nonconv}, we restrict to situations
%with convex obstacles {\it having exactly one top}. The non-generic
%case of convex obstacles with a horizontal line of tops is not
%considered.
Related to the geometrical description of the obstacles is the
concept of {\it top of an obstacle}, which plays an important role
in the construction of the coverage algorithm and its analysis.
%Although the major part of this paper considers convex obstacles,
%\secref{sec:nonconv} treats nonconvex generalizations. Therefore a
%general definition is given.
\begin{defi}
\label{def:top} A top $p=(p_x,p_y) \in \set{R}^2$ of a (convex)
obstacle $P$ is a point satisfying the following property:
\[\exists \delta > 0, \forall \epsilon < \delta : q=(q_x,q_y)
\in N(p, \epsilon) \cap P \Rightarrow  q_y \leq p_y,\] where
\[N(p, \epsilon) := \{z \in \set{R}^2 | \Vert z-p \Vert < \epsilon\}.\]
\end{defi}
The ``top'' is defined relative to the direction of motion of the
robot group: \defref{def:top} assumes a robot group moving parallel
to the y-axis towards larger $y$-values. If the robot group moves in
the opposite direction, the definition changes correspondingly by
demanding $q_y \geq p_y$. A convex obstacle either has exactly one
top or possesses a connected line of tops.

\subsection{Robot sensors and communication}
\label{sec:set}

Each robot is equipped with two types of omnidirectional sensors.
One type serves as a means to detect the targets in the assigned
area, e.g. landmines. Its detection range is denoted $r^+_t$. The
other type is responsible for detecting obstacles and other robots,
with detection range $r^+_r$. The simplest sensor model available is
the binary detection model: a target is detected (not detected) with
complete certainty if it is in inside (outside) the sensor's
detection range \cite{Sundaraja}. More realistic descriptions of a
sensor's detection capability use probabilistic models \cite{Elfes}.
A function $c_p : \set{R}^2 \to [0,1]$ expresses the {\it coverage
confidence level}: the value $c_p(q)$ represents the probability
that the sensor located at $p$ detects an object that is located at
$q$. We assume the confidence level only depends on the distance
between $p$ and $q$, which leads to a circular symmetry around the
sensor. We further assume the confidence level to be defined as
follows:
\begin{equation}
c_{p}(q) =
\begin{cases}
1, &  || p - q || \leq r^-,\\
\frac{e^{- || p - q ||} - e^{-r^+}}{e^{-r^-} - e^{-r^+}}, &   r^- < || p - q || < r^+,\\
0 , &  || p - q || \geq r^+.
\end{cases}
\label{eq:probsensmodel}
\end{equation}
In a circular area around the sensor (with radius~$r^- \in
\set{R}_{> 0}$) targets are always detected. this area is surrounded
by a ring-shaped area (with radius inside $(r^-,r^+)$) where targets
are detected with a probability smaller than $1$. Outside the area
with radius $r^+$ targets are never detected. We use
(\ref{eq:probsensmodel}) to describe the sensor capabilities, by
adding the subscript $r$ for sensors detecting other robots or
obstacles and $t$ for sensors detecting the targets specified in the
robot coverage mission.

Furthermore, each robot possesses a compass enabling the robot to
determine its orientation within space. The compass is used to align
all robots along the same absolute reference direction, parallel to
the strip boundary. The two outer robots of the group (robots $1$
and $6$ in \figref{fig:init}) are equipped with a GPS system to
determine their position as explained in the introduction.

Communication is limited: first, it is only allowed between robots
that sense each other. Two robots sense each other if and only if
they are sufficiently close to each other (expressed by the maximum
detection range $r^+_r$) and if there is no obstacle located on the
straight line connecting them (so-called line-of-sight
communication). Second, each robot transmits a {\it limited} set of
messages, concerning its status, with the purpose of inducing a
change of behavior in other robots.

\subsection{Robot formation}
\label{sec:robfor}

Consider a population of $N$ robots. To every robot one assigns an
index number $i \in \{1,\ldots,N\}$, which serves as the robot's
identity. Each robot $i$ with $2 \leq i \leq N$ has an {\it
Immediate Leader} (denoted IL) with index $i-1$. Similarly each
robot $i$ with $1 \leq i \leq N-1$ has an {\it Immediate Follower}
(IF), with index $i+1$. The position of robot $i$ is given by
$q_i:=(x_i,y_i) \in S$. We assume holonomic robots with discrete
dynamics
\begin{equation}
q_i[k+1] = q_i[k] + u_i[k], \quad 1 \leq i \leq N, \: k \in \set{N},
\end{equation}
where $u_i \in \set{R}^2$ is the control input to the $i$th robot.
The input is bounded: $||u_i(k)|| \leq \tilde{v}, \forall i,k$,
where $\tilde{v}$ is the distance traveled during one unit of time
when moving at the maximum allowed velocity $v_{\mathrm{max}}$.

\begin{figure}[ht]
\centering
\includegraphics[width=5cm]{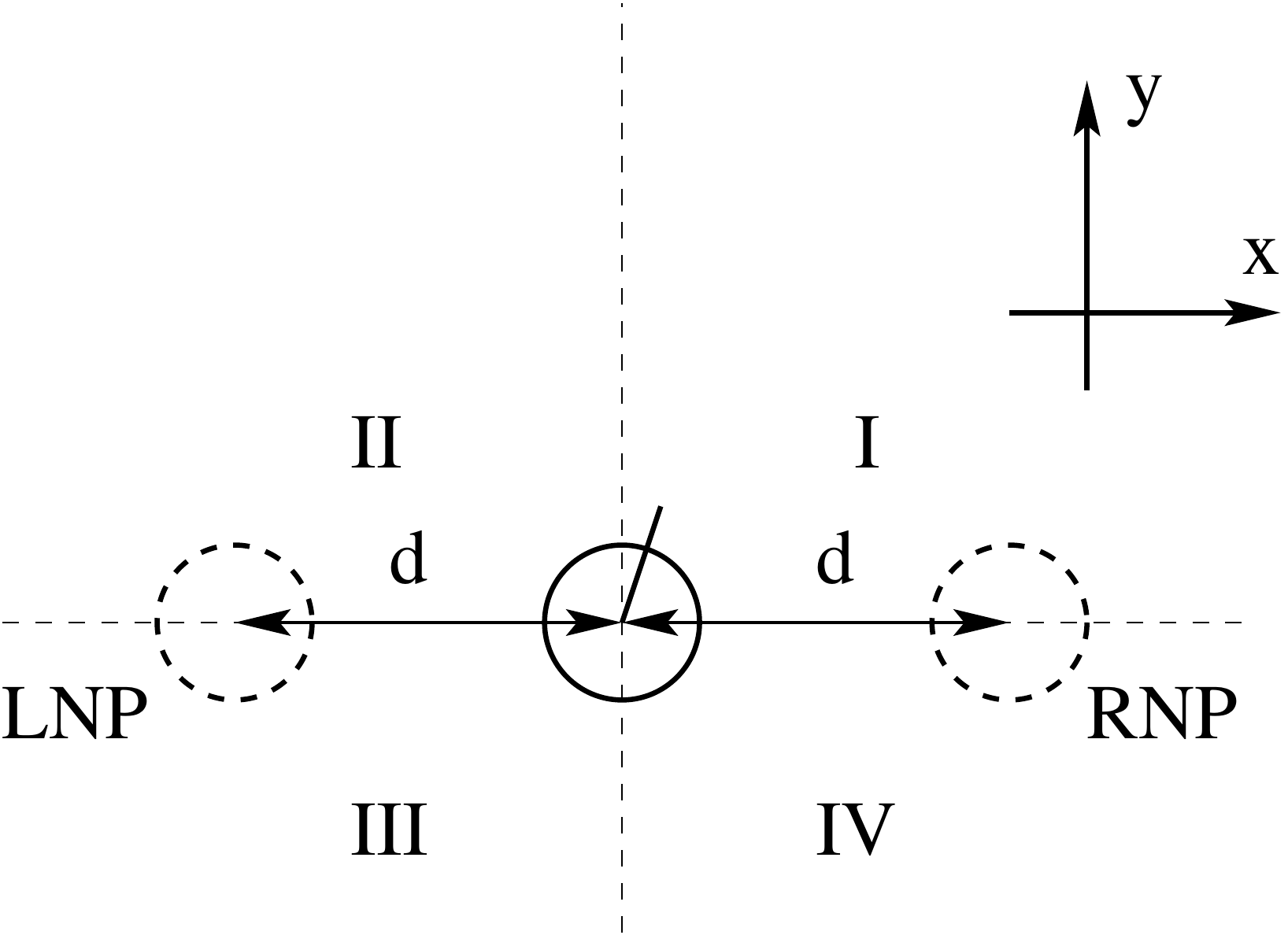}
\caption{Schematic representation of the scanning of consecutive
scanning strips.} \label{fig:kwadrant}
\end{figure}

\begin{figure}[ht]
\centering
\includegraphics[width=8cm]{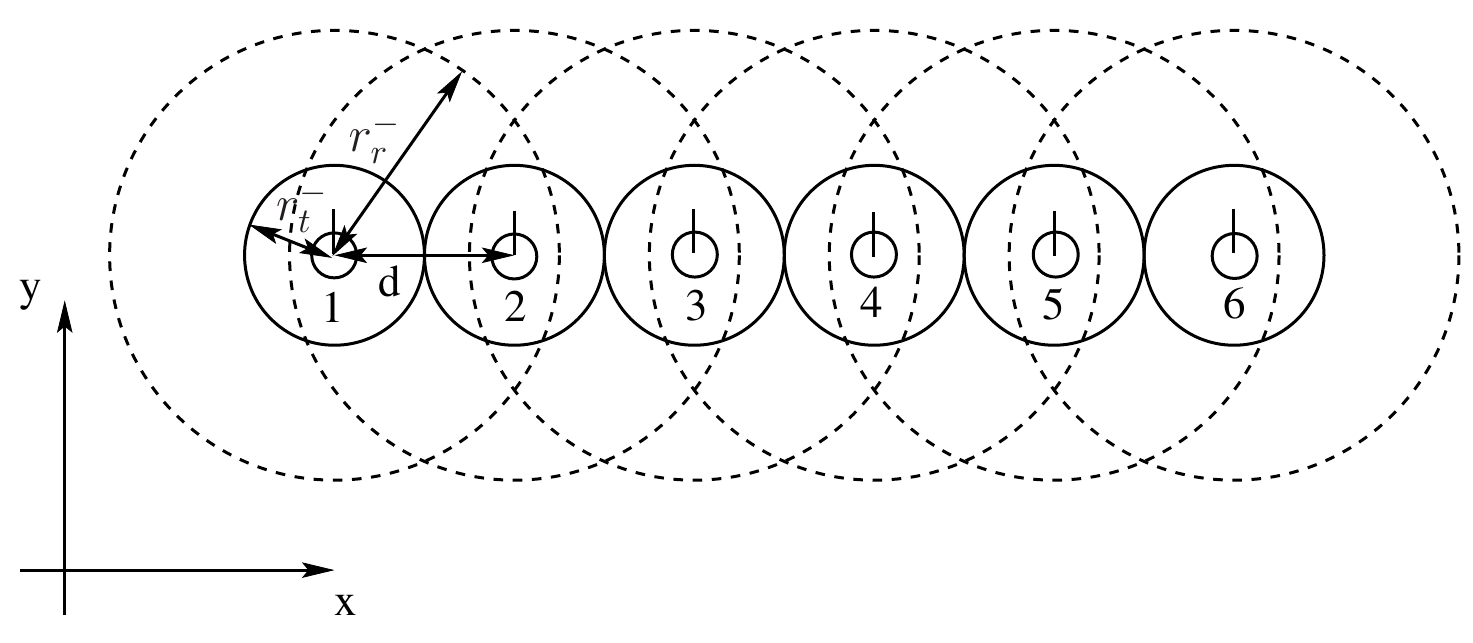}
\caption{A desired robot configuration of $6$ robots. The small
solid circles represent the robots; the large solid circles show the
area where targets are detected with probability equal to one; the
dashed circles indicate the area where robots are able to detect
each other with probability equal to one.} \label{fig:init}
\end{figure}

%%Licht wijzigen: orientatie van de robot is niet belangrijk
The robot formation used throughout the algorithm is defined by
relative positions among the robots.
%defined as follows. The forward direction of each robot is oriented
%along the positive y-axis.
%This gives meaning to the concepts
%``left'' and ``right'' and
We introduce the following definition (depicted in
\figref{fig:kwadrant}).
\begin{defi}\label{def:LNP}
With $d \in \set{R}_{>0}$, the Left Neighbor Position (LNP) of a
robot at $(x,y)$ is the point with coordinates $(x-d,y)$; similarly
the Right Neighbor Position (RNP) is the point $(x+d,y)$.
\end{defi}
The robot formation is obtained when the IL of each robot is located
at this robot's LNP. This yield a straight-line robot formation
oriented perpendicular to the direction of motion along the y-axis,
as shown in \figref{fig:init} for $N=6$. The solid circles delimit
the area sensed for targets by each robot with coverage confidence
level equal to one; they have radius $r^-_t$. These circles overlap
if the inter-robot distance $d$
satisfies $d \leq 2 r^-_t$. %These coverage confidence levels of
%neighboring robots to overlap. The overlap is such that their
%circular areas with confidence level equal to one touch each other.
We set $d = 2 r^-_t$ such that along the line connecting two
neighboring robots in formation, targets are detected with
certainty.
%Overlapping sensor ranges improve the algorithm's performance in a
%second way. Eq. (\ref{eq:probsensmodel}), modeling sensor accuracy,
%includes a ring-shaped area around the sensor where target detection
%is not guaranteed. When only covered by this ring-shaped area, some
%targets could get passed unseen. The proposed robot formation is
%such that neighboring robots partially scan the same area,
%increasing the detection probability.
%restriction is necessary to retrieve moving targets as motivated in
%\secref{sec:run2stand}.
%the next
%section where the%To ensure partially overlapping target sensor
%%ranges of neighboring robots, $d/2 < s_t$ needs to be satisfied.
%algorithm is described. %% Iets preciezer zijn?
%Naturally, it is assumed that $s_t < s_r$. The larger $s_t$, the
%farther we can set neighboring robots apart.
%This restricts the maximum inter-robot distance: $\frac{d}{2} <
%r^{-}_{t}$
Similarly, it is demanded that neighboring robots detect each other
with a probability equal to one. Therefore we require $d \leq
r^-_r$. In \figref{fig:init} this corresponds with overlapping
dashed circles with a robot located in each intersection.

% the
%inter-robot distance $d$ is required to be smaller than or equal to
%the $r^-_r$-value of the corresponding sensors.

\paragraph*{Remark}
Notice that \figref{fig:init} indicates inter-robot distances as the
distance between the robots' centers. This is the definition of
inter-robot distance used throughout the paper. It allows us to
neglect the real size of the robots in a theoretical approach of the
scanning algorithm. In practice however, the distance sensors of the
robot are attached to the robot's outer frame. A version of the
algorithm used in practice needs appropriate adjustments to take the
robot size into account.

\subsection{Problem statement}
\label{sec:prob}
\begin{figure}[ht]
\centering
\includegraphics[width=5cm]{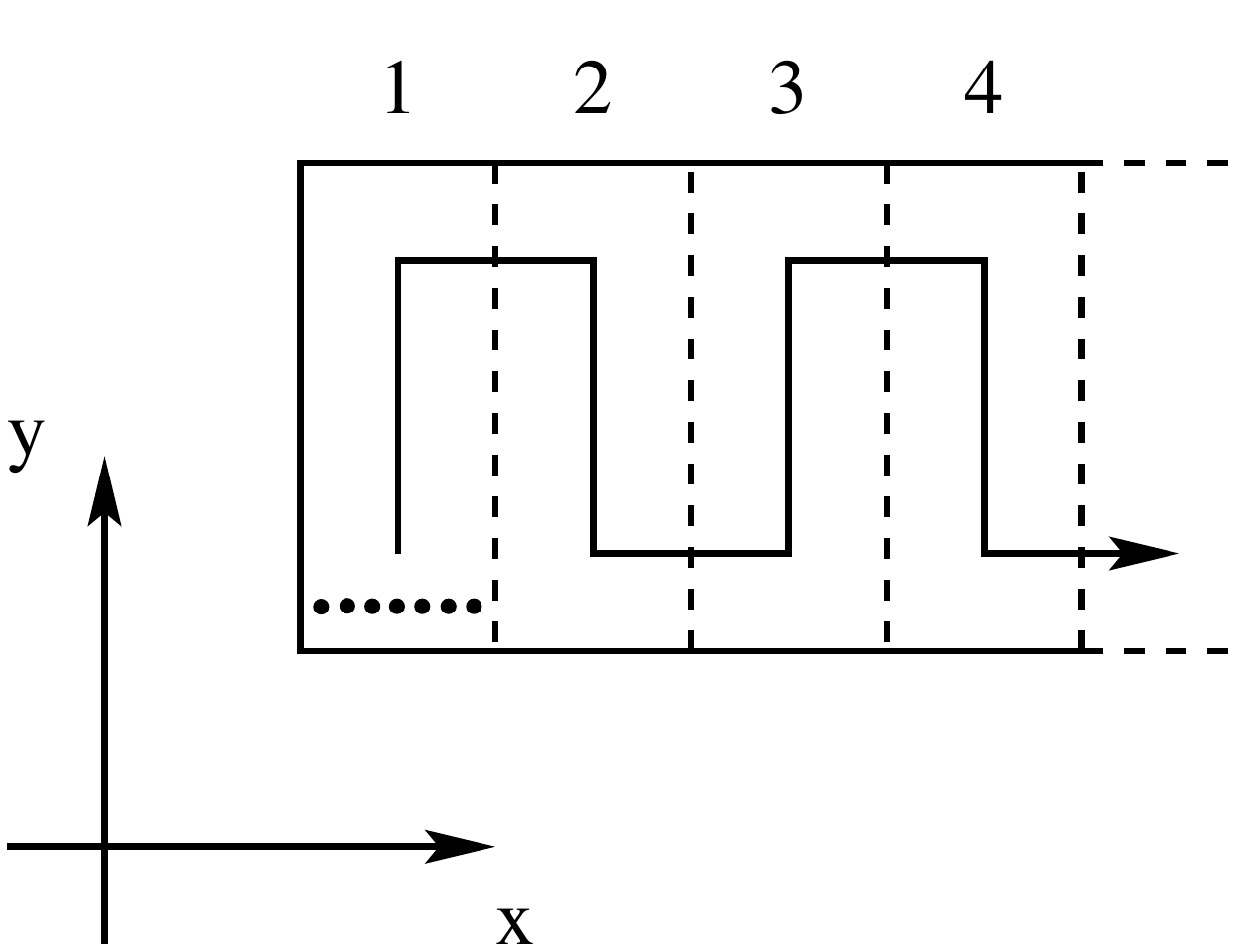}
\caption{Schematic representation of the scanning of consecutive
scanning strips.} \label{fig:striptrans}
\end{figure}

%\subsubsection{Locating fixed targets}
Call $\mathcal{P}$ the set of all points of $S$ belonging to
obstacles: $\mathcal{P} := \{q | q \in P_{i_O}, i_O \in
\mathcal{N}_O\}$, with $P_{i_O}$ an obstacle. In order to locate all
fixed targets, we need the robots to cover $S \setminus
\mathcal{P}$. We demand the robots cooperate with each other by
maintaining the formation defined in \secref{sec:robfor}. The robot
formation performs a sweep of each scanning strip. The consecutive
strips are scanned in opposite directions as illustrated in
\figref{fig:striptrans}. The challenge is to pass all obstacles,
with unknown size and location, so that coverage is still
guaranteed. The robot group is allowed to split into subgroups to
move past obstacles. The shape of the robot formation, i.e. the
straight line, will allow for easy and accurate reconnection of
subgroups.

\paragraph*{Remark}
The algorithm alternates between stages where the robot group
advances towards larger y-values and stages where it moves towards
lower y-values. In \secref{sec:alg}, the algorithm is explained and
described for the former stages, where the above \defref{def:LNP}
for LNP and RNP is valid. In the latter stages, where the robot
group has reversed its traveling direction, the LNP and RNP can be
redefined by simply substituting $-d$ for $d$.

%\subsubsection{Locating moving targets}
%\label{sec:movtarget} In this problem setting, the targets are
%allowed to move freely in $S \setminus \mathcal{P}$, but cannot exit
%this area. With the additional assumption that every moving target
%is unable to exit the scanning strip it is in, we keep the scanning
%procedure of \figref{fig:striptrans}. The challenge is to pass all
%obstacles in $S$ without leaving unsensed gaps in between robots or
%between robots and obstacles through which the moving targets could
%pass unseen. At the end of the sweep all moving targets have been
%intercepted. If the additional assumption cannot be maintained, we
%consider the area $S$ as one big scanning strip. We make the robot
%group sufficiently large so it can scan $S$ in one sweep.
%%Since we restrict the moving targets to stay inside
%%$S$, it is reasonable to consider the boundary of $S$ as a solid
%%wall. The algorithm developed in the next section is restricted to
%%obstacles that do not touch this solid boundary.

\section{Preliminaries to the scanning algorithm}

%In this section some preliminary definitions and the different
%parameters constituting the state of a robot are introduced.

%The second section describes each behavior type in detail, and
%describes the algorithm with the aid of an example.
%
%The third section gives a proof of coverage and shows the robot
%formation never ends up in dead-lock situations.

\subsection{Preliminary definitions}
%This initial assignment of leader-follower pairs can change during
%the task performance, due to the presence of obstacles, see further
%in \secref{sec:strip}. Robot~$i$ maintains a constant distance $d<
%s_r$ with its IL and observes its IL at an angle of $\pi/2$ with
%respect to its forward direction.

\begin{defi}
A location in space is called {\it covered} at time~$t_1$ if it was
found inside the sensor range $r^-_t$ of one of the robots at some
time instant $t \leq t_1$.
\end{defi}

\begin{defi}
A location in space is called {\it obstructed} to a robot if the
straight line connecting that location with the present position of
the robot intersects an obstacle or other robot.
\end{defi}

%\begin{defi}
%A location in space is called {\it distant} to a robot if it is not
%obstructed and the robot is unable to reach it at the preset
%velocity $v$ in one time step.
%\end{defi}

\begin{defi}
A location in space is called {\it reachable} to a robot if it is
not obstructed and the robot is able to reach it at the preset
velocity $v$ in one time step.
\end{defi}

\begin{defi}
The robots are classified into three groups according to their
location in the formation. The leftmost and rightmost robots are
called the {\it Left Strip Boundary robot}, denoted LSB, and the
{\it Right Strip Boundary robot} (RSB). The third group consists of
all remaining robots; they are called {\it Interior Robots} (IR).
\end{defi}

With the $(x,y)$-frame defined in \secref{sec:modenv} we are able to
define positions w.r.t. obstacles. With $y_{min}$ and $y_{max}$
defined as the smallest resp. largest y-value of an obstacle $P$ we
define
\begin{enumerate}
    \item ``on the left of $P$'' $:=$ located at one or more points $(x_i,y_i)$
     with $y_i \in (y_{min},y_{max})$ and $x_i \not \in P$ such that $\exists x\in $P$ : x_i <
     x$,
    \item ``on the right of $P$'' $:=$ located at one or more points $(x_i,y_i)$
     with $y_i \in (y_{min},y_{max})$ and $x_i \not \in P$ such that $\exists x\in $P$ : x_i
     > x$.
\end{enumerate}

\subsection{Necessary sensor data}
Since every robot is equipped with a compass it is able to retrieve
the orientation of the (x,y)-frame defined at initialization. Each
robot divides its surroundings into {\it four quadrants} with
respect to the frame that
\begin{itemize}
    \item is a translated version of the initial (x,y)-frame and
    \item has the center of the robot located at the origin.
\end{itemize}
%Besides the introduction of quadrants, this frame of axis enables us
%to define relative positions as follows:
%\begin{enumerate}
%    \item forward: at larger y-values, but with the same x-value,
%    \item left: at smaller x-values, but with the same y-value,
%    \item right: at larger x-values, but with the same y-value.
%\end{enumerate}
These quadrants are denoted by Roman numerals in
\figref{fig:kwadrant}. Furthermore, every robot is equipped with
sensors measuring both
\begin{itemize}
    \item the distance between itself and nearby obstacles and robots,
    \item  the direction along which these obstacles and robots are detected. The straight
half-line along this direction belongs to one of the robot's four
quadrants.
\end{itemize}
In the case of sensed obstacles, the robot stores into its memory
\begin{itemize}
\item the shortest distance between itself and the surrounding
obstacles, denoted Dist2O (shorthand for ``Distance w.r.t.
Obstacles''),
\item the quadrant corresponding to this shortest distance.
\end{itemize}
In the case of sensed robots, the robot stores into its memory all
inter-robot distances and corresponding directions.

\subsection{A robot's parameters}\label{sec:robpar}
Besides its position in $S$, the state of every robot is determined
by a number of parameters.

\subsubsection{Behavior type (TYPE)} The TYPE parameter assumes one
of three values: Run-Mode, Contour-Following, and Standstill:
% Their general meaning is as follows. (A
%more precise description of these behavior types is given in the
%next section.)
\begin{itemize}
    \item Standstill: do not move.
    \item Contour-Following:
    move clockwise along the boundary of the nearest obstacle.
    \item Run-Mode: If the IL of a robot $i$ is not within sensor range
     or it is Contour-Following, robot $i$ moves forward with preset velocity~$v$;
    otherwise, i.e. if the IL is located at $(x_i,y_i)$ within sensor range
    and is not Contour-Following, robot $i$ moves to $(x_i + d, y_i)$, called the Desired
    Position (DP).
\end{itemize}

\subsubsection{``Located at the top of an obstacle (TOP)''} In some
instances it is necessary to mark the robot as being located at the
top of an obstacle. The parameter values of TOP are ``on'' and
``off''.

\subsubsection{``Located left of an obstacle (LEFT)''} This parameter
indicates when a Standstill robot (group) is located on the left of
the nearest obstacle. The parameter values of LEFT are ``on'' and
``off''.

\subsubsection{``Located right of an obstacle (RIGHT)''} This parameter
indicates when a Standstill robot (group) is located on the right of
the nearest obstacle. The parameter values of RIGHT are ``on'' and
``off''.

%P2O is related to the position of the robot with respect to the
%obstacles. It can assume the values Left, Right, or Undetermined.
%The meaning and function of these P2O-values is explained in
%\secref{sec:strip}, where it is described how a robot changes from
%Run-Mode to Stand-Still. Every robot communicates the couple
%(TYPE,P2O) to its IL and IF (if the communication-enabling
%conditions are fulfilled). At the same time it is receiving this
%information from its IL and IF, and responds appropriately to
%changes in value.
% Once the above preferred formation is attained the
%scanning algorithm is initialized. The robot group will cover all
%free space of area $S$ by sweeping strip after strip in a
%zigzag-like pattern.
% Stated more precisely, we demand that the inter-obstacle
%distance measured perpendicular to the strip boundary be larger than
%the inter-robot distance $d$.
\begin{figure}[t]
\centering
\includegraphics[width=3.5cm]{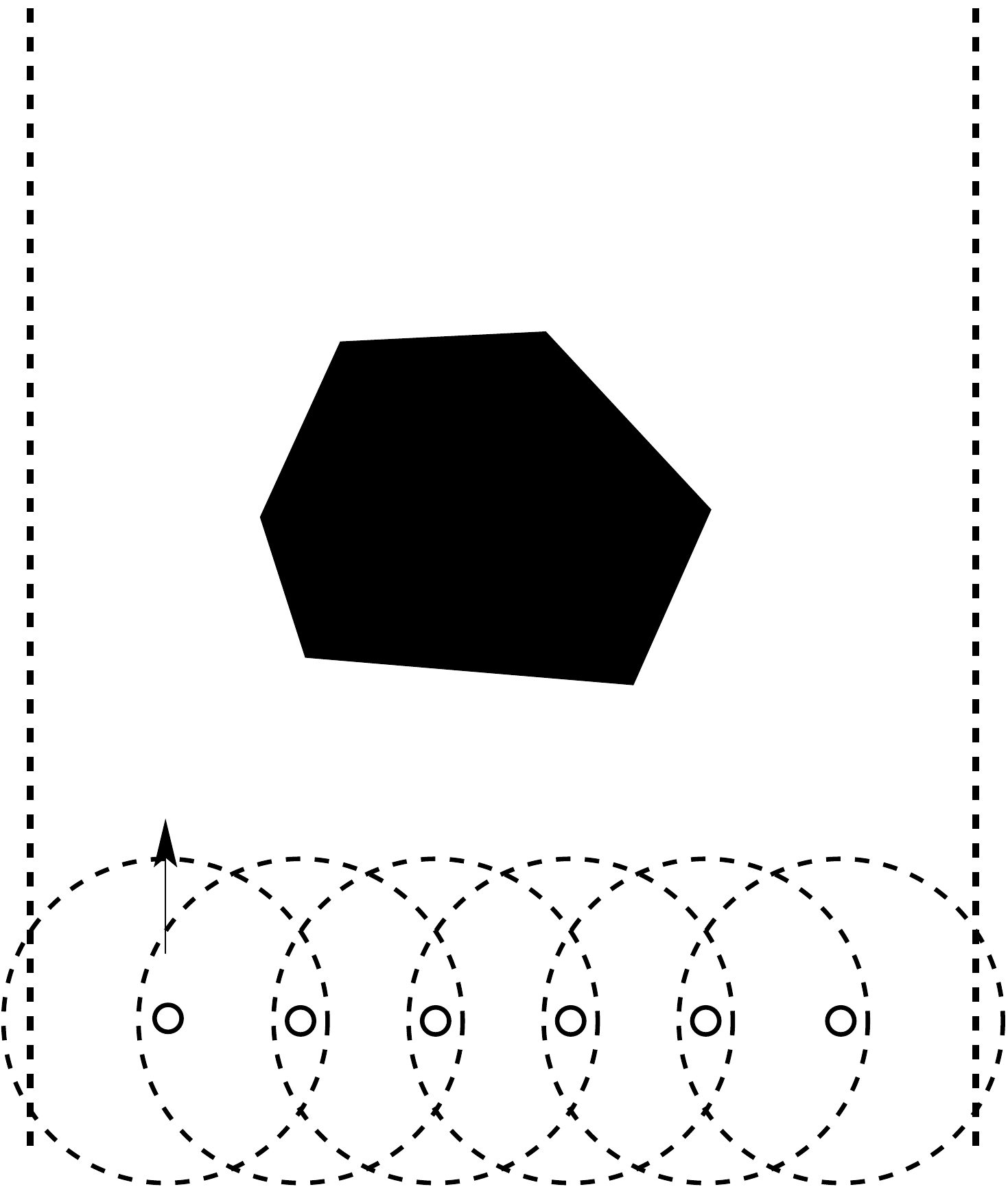}
\caption{A depiction of the algorithm.} \label{fig:algo}
\end{figure}\begin{figure*}[ht]
\centering
  \subfloat{\includegraphics[height=3cm,clip]{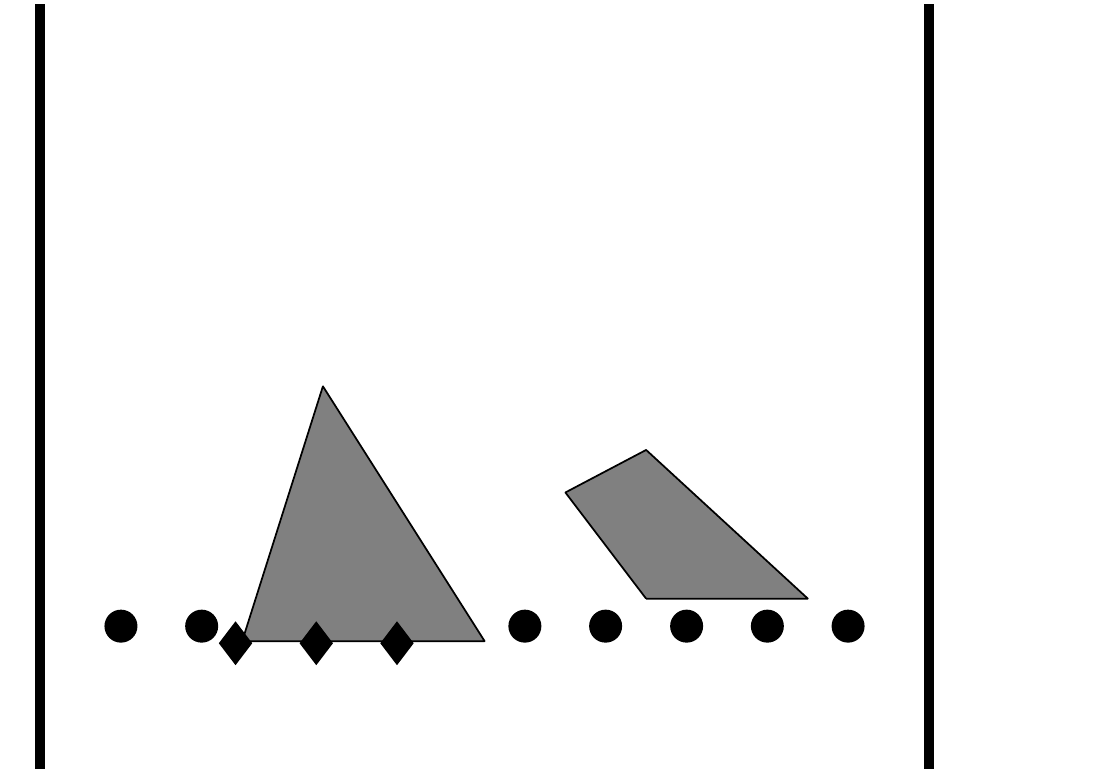}} \quad
  \subfloat{\includegraphics[height=3cm,clip]{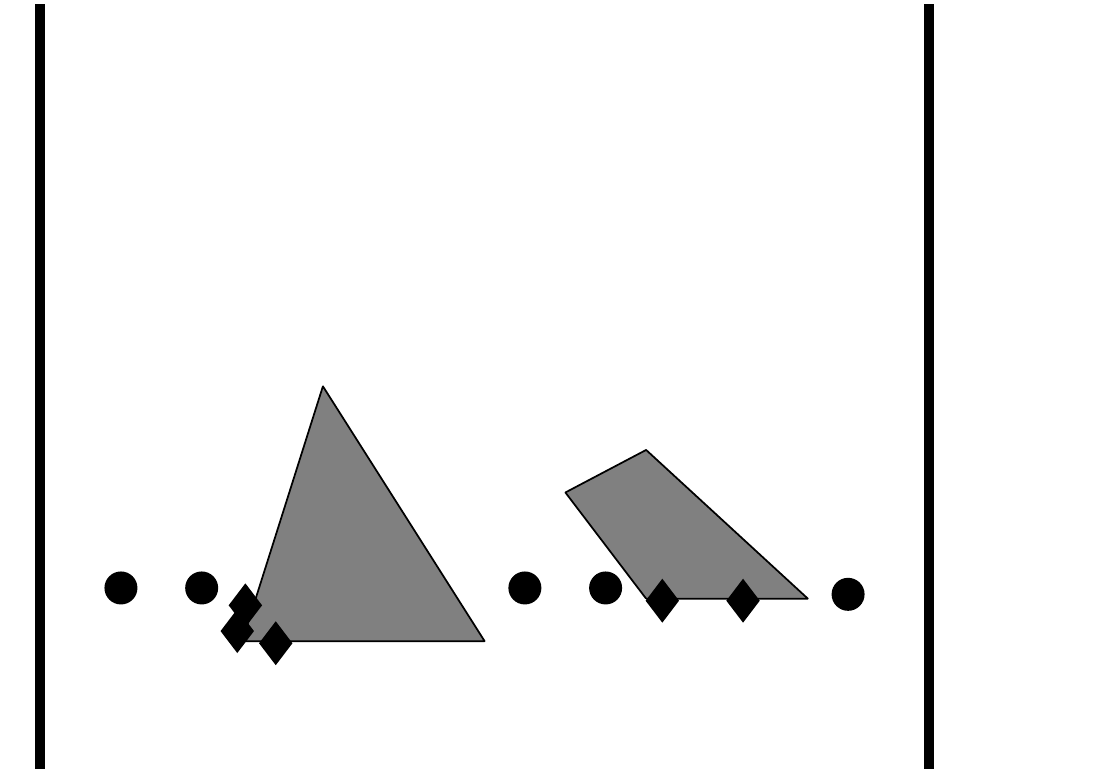}} \quad
\subfloat{\includegraphics[height=3cm,clip]{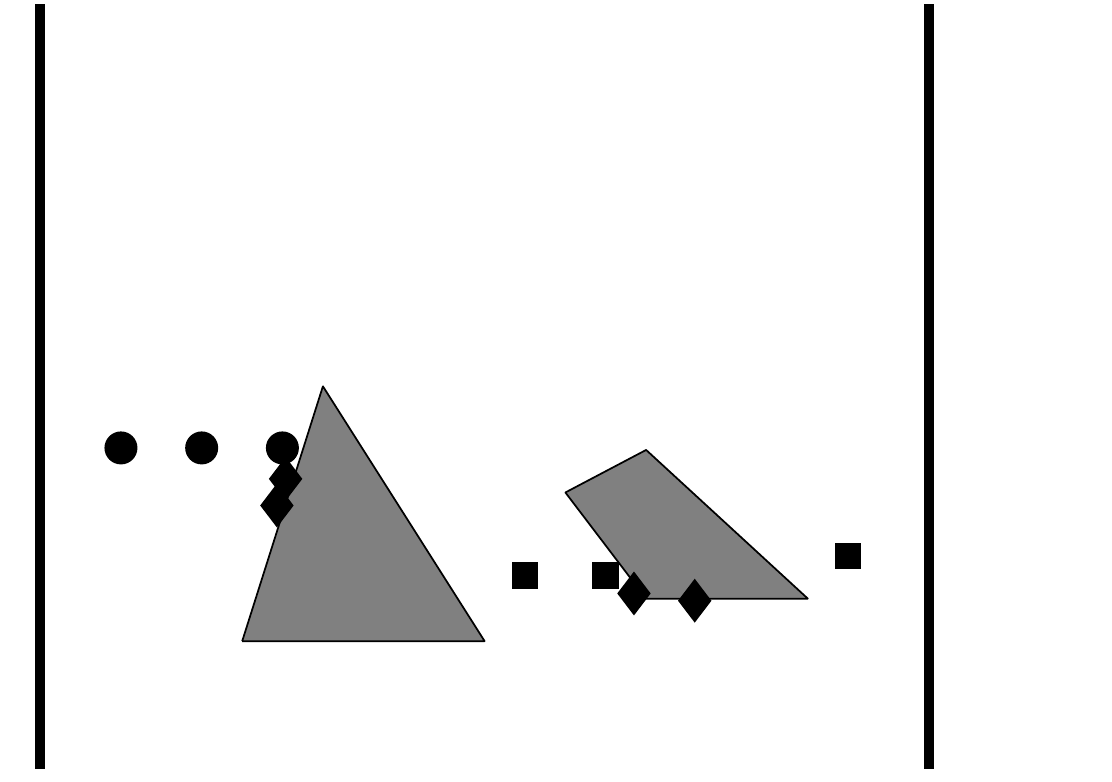}} \quad
\subfloat{\includegraphics[height=3cm,clip]{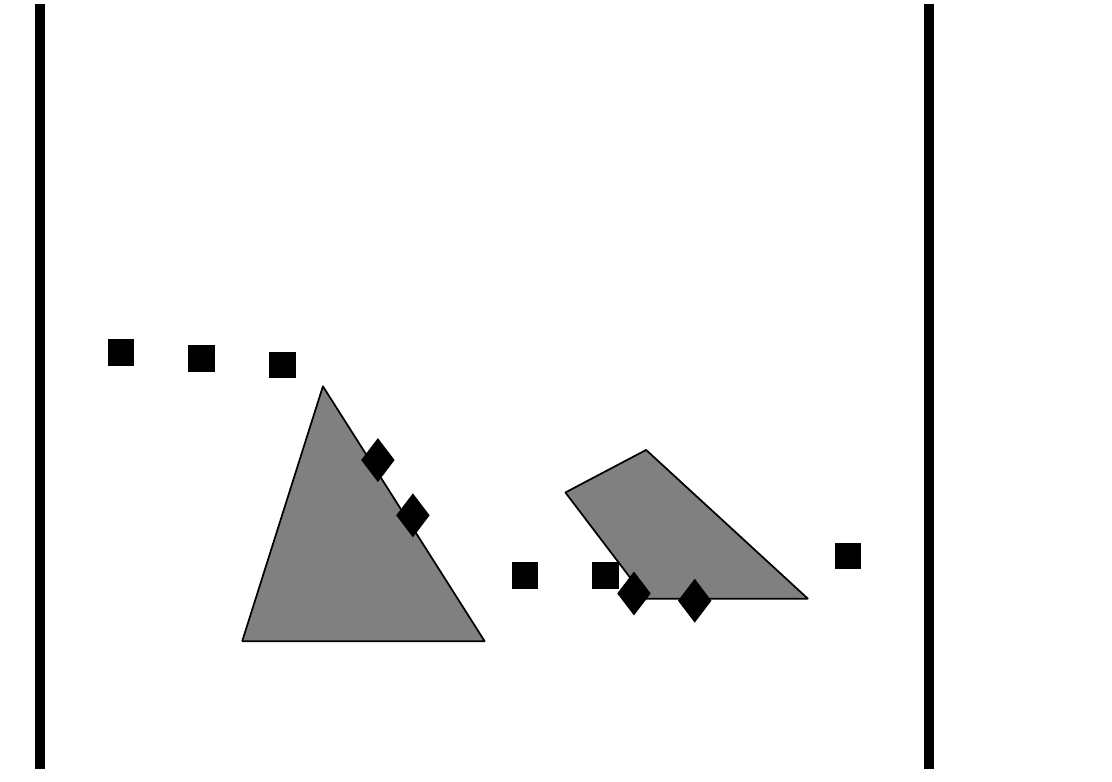}} \\
\vspace{.5cm}
 \subfloat{\includegraphics[height=3cm,clip]{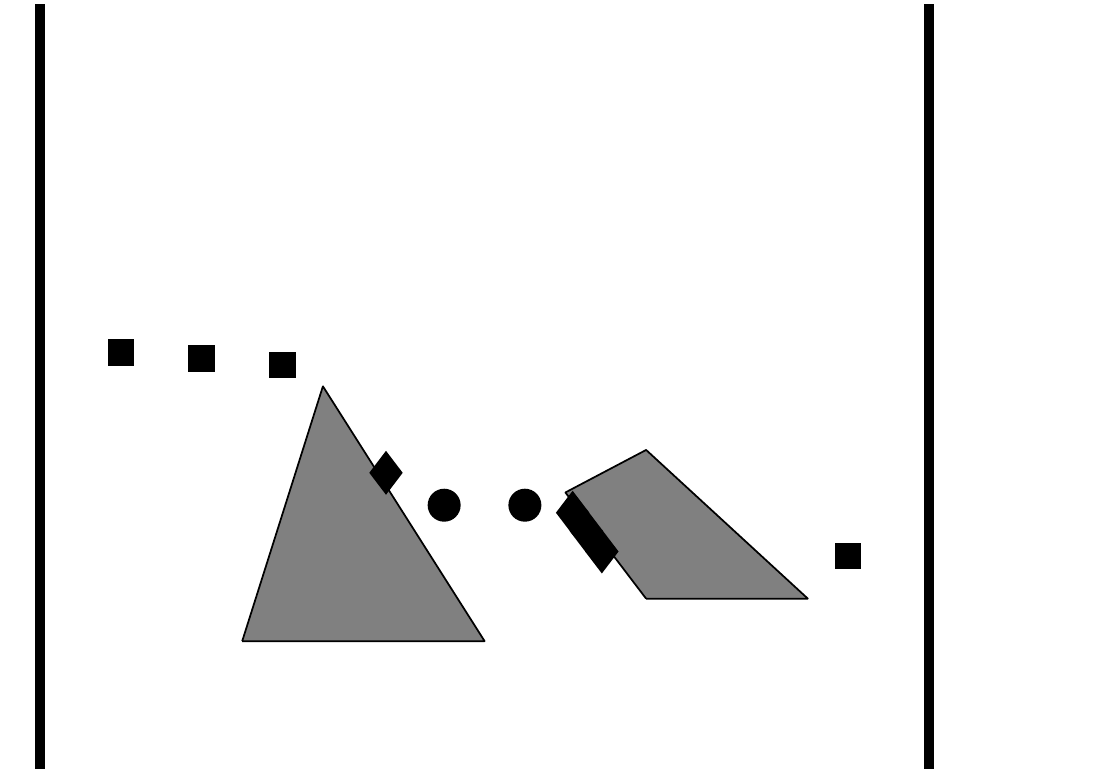}} \quad
 \subfloat{\includegraphics[height=3cm,clip]{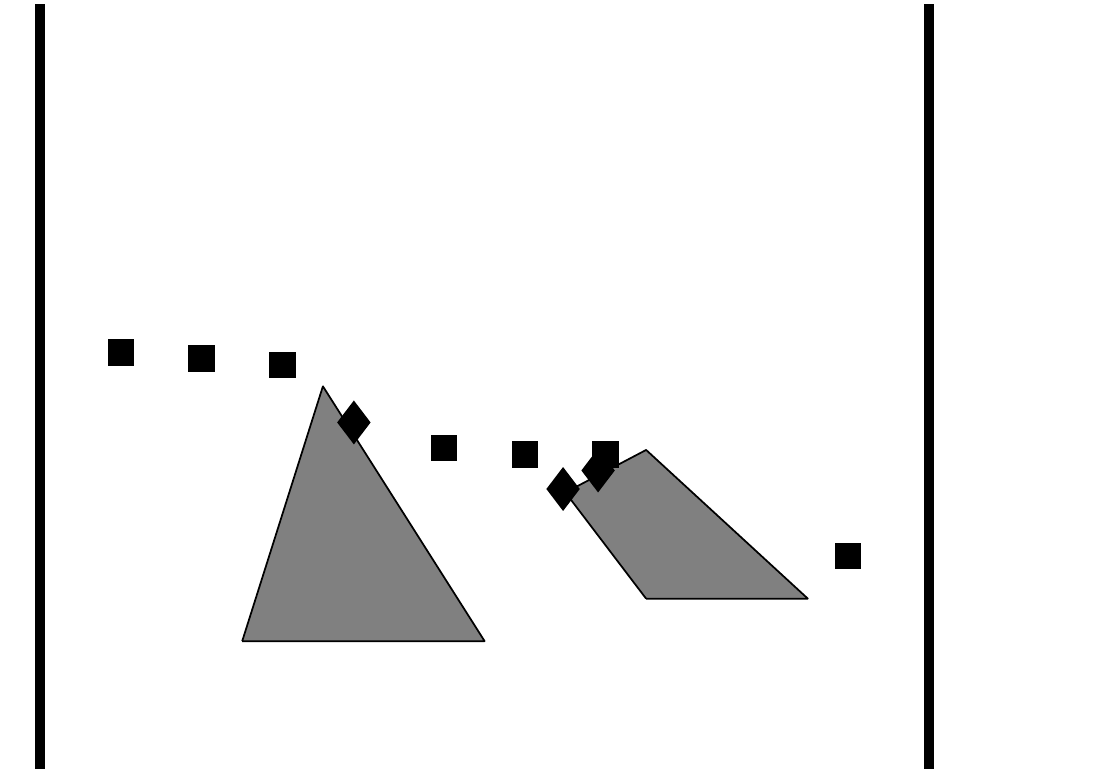}} \quad
  \subfloat{\includegraphics[height=3cm,clip]{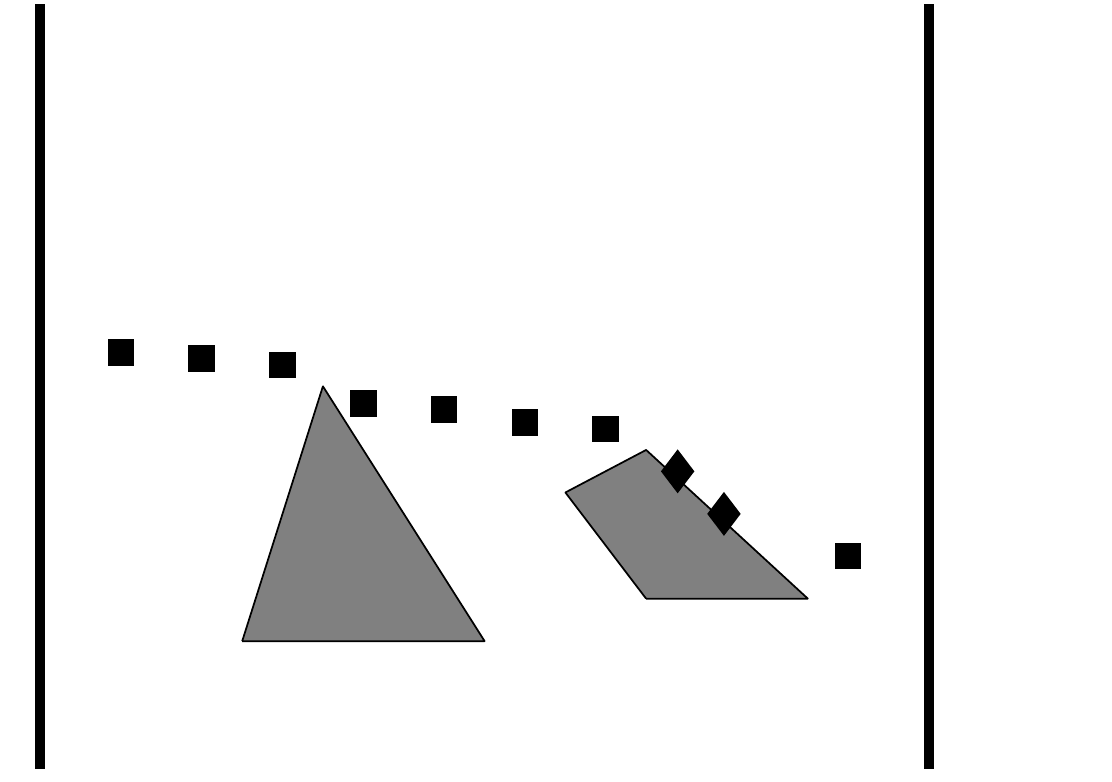}} \quad
  \subfloat{\includegraphics[height=3cm,clip]{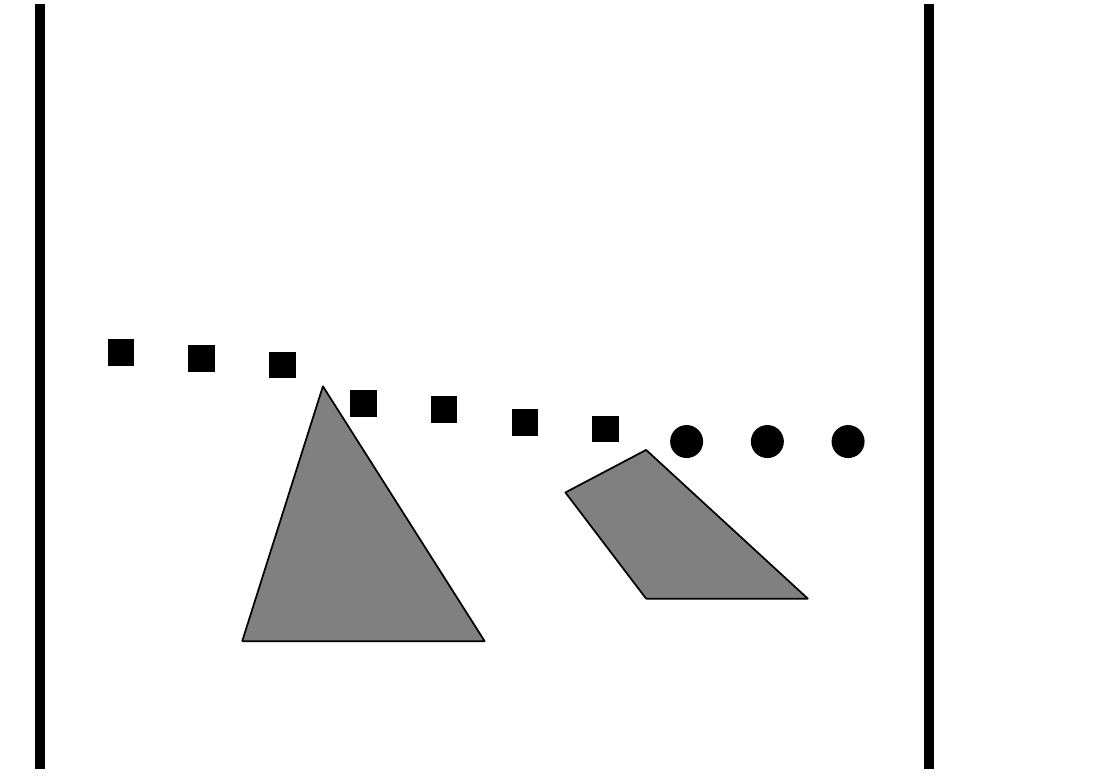}}
  \caption{A 10-robot group passing two disjoint obstacles simultaneously. Squares $=$ robots at Standstill;
  diamonds $=$ Contour-Following robots; disks $=$ robots in Run-Mode.}\label{fig:obstsim}
\end{figure*}
\section{Scanning one strip without obstacles}\label{sec:strip}
Consider a scanning strip devoid of obstacles. The width of the
scanning strip $w$ and the number of robots in the formation $N$ are
interdependent. From the settings in \secref{sec:robfor}, the
formation width equals $(N-1)d$. Referring to
(\ref{eq:probsensmodel}), the distance between each outer robot and
its respective strip boundary is set to $r^-_t$, to ensure detection
of targets located near the strip boundary. This yields a
corresponding strip width
\begin{equation}
w = (N-1)d +2 r^-_t. \label{eq:wN}
\end{equation}
Given a value for $w$, (\ref{eq:wN}) yields the necessary number of
robots to scan the strip in one sweep.

At the initialization of the algorithm, the robot group has assumed
the formation defined in \secref{sec:robfor}. With $(x_1,y_1)$ the
coordinates of LSB, the horizontal position of the $i$th robot is
$x_i=x_1 + (i-1)d$. The horizontal line $y=y_1$ is covered by the
robot sensors for all $x \in [x_1- r^-_t,x_N + r^-_t]$. Robot LSB
executes its algorithm which consists of tracing a straight line
$x=x_1$, $y \geq y_1$ at a constant velocity $v$. The remaining
robots maintain the formation and move along their respective
straight line $x=x_i$, $y \geq y_1$. When the LSB stops at
$(x_1,y_2)$ the corresponding area $A := [x_1- r^-_t,x_N + r^-_t]
\times [y_1,y_2] $
%\[A_{(x_1,x_N,y_1,y_2)} := \{(x,y) | x \in  \}\]
has been completely covered by the sensors. In other words, if the
robots track the parallel lines with x-coordinates $x_1 +id,
i=1,\ldots,N $, the entire area gets covered.

%We obtain the following proposition.
%\begin{prop}
%\label{prop:1} Consider a run of the algorithm starting at time
%$t_0$ and finishing at $t_1$. If each point $q:=(q_x,q_y)$ belonging
%to the set of lines
%\[V:=\{ q | q_x=,\: q_y \in [y_1,y_2]\}\]
%is covered by the center of a robot at a time instant $t \in
%[t_0,t_1]$, then the corresponding area $A_{(x_1,x_N,y_1,y_2)}$ has
%been completely covered by the robot sensors at time $t_1$.
%\end{prop}

When the robot group reaches the end of a strip, it moves to the
start of the next strip. All robots turn $180$ degrees, the LSB and
RSB exchange roles, and the robot team commences a new sweep.

% . The algorithm needs to be
%extended, preferentially with as few extra routines as possible. The
%strips are scanned in a strict order in a back-and-forth fashion.
%Each strip receives an index number. Transitions from odd to even
%numbered strip differ from those from even to odd (see
%\figref{fig:striptrans}).
%
%Between consecutive strips,  as follows. Both robots keep track of
%the index number of the scanning strip under consideration. After
%each strip transition this number is augmented by one. When the
%robot group reaches the end of the strip, the current LSB checks the
%index number of the strip. If it is odd, the LSB moves one strip
%width to the right, making use of its GPS equipment; if the number
%is even, the robot goes one strip width to the left. As programmed,
%the other robots follow the LSB so that the entire group ends up at
%the start of the next strip. Then, the LSB sends the following
%command to its IF:
%\begin{itemize}
%\item align to the originally prescribed direction rotated over 180 degrees,
%\item turn your couple (IL,IF) into (IF,IL).
%\item Send this command to your newly assigned IL,
%\end{itemize}
%In the meantime the LSB rotates itself 180 degrees, and takes on the
%role of the RSB. The original RSB of the group is programmed to
%interpret the third part of the above command as taking on the role
%of LSB. The robot group is now ready to start scanning the next
%strip.

\section{Scanning one strip with obstacles}
\label{sec:alg} Before presenting a detailed description of our
algorithm, we give a brief sketch of the main idea. Every robot of
the formation moves along its predefined straight line, until the
line is obstructed by an obstacle, in which case the robot moves
clockwise around the obstacle. Subgroups of robots which are not
hindered by the obstacle keep advancing past obstacles up to a
maximum distance from the obstacle. These subgroups then wait for
robots moving along the obstacle to (re)join, after which the
enlarged subgroup is able to advance further. In this way subgroups
build up alongside the obstacle until eventually two subgroups from
each side of the obstacle join at the obstacle's top. This idea is
depicted in \figref{fig:obstsim}, where two obstacles are being
passed simultaneously. The figure is the result of a computer
simulation implementing the algorithm as it is described below.

In more detail, the solution to the problem statement of
\secref{sec:prob} consists of endowing each robot with the behavior
types Standstill, Run-Mode, and Contour-Following defined in
\secref{sec:robpar}. These behavior types are each triggered by
obstacles and/or other robots. The algorithm itself consists of a
set of rules that determine the conditions forcing appropriate
transitions between the three behavior types. These rules are
described in $3$ tables, one for each behavior type (see
\algoref{alg:Run-Mode}, \algoref{alg:Contour-Following}, and
\algoref{alg:red}). As will be explained in the present section,
they lead to successful coverage if the minimum inter-obstacle
distance is given by $2 \sqrt{2}d$. %The interaction between robots
%using these switching rules is illustrated by

\begin{algorithm}
\caption{TYPE Run-Mode} \label{alg:Run-Mode}
    \begin{algorithmic}[5]
        \State Repeat
        \If {IL is visible $\land$ its TYPE is not Contour-Following}
        \State  {\it keep your IL at your LNP}
        \Else \State {\it Advance with constant $x$-coordinate}
        \EndIf\\
        until
        \If{ An obstacle is blocking the forward path}
            \State TYPE $\pijl$ Contour-Following
        \ElsIf{an obstacle in quadrant 4 $\land$ dist2O $> d$ $\land$ your IF is not at your RNP}
                \State LEFT $\pijl$ ON
                \State TYPE $\pijl$ Standstill
        \ElsIf{ your IF is at your RNP, with TYPE(IF)=Standstill and LEFT(IF)=ON}
                \State LEFT $\pijl$ ON
                \State TYPE $\pijl$ Standstill
        \ElsIf{an obstacle in quadrant 3 $\land$ dist2O $> d$ $\land$ your IL is not at your LNP}
                \State RIGHT $\pijl$ ON
                \State TYPE $\pijl$  Standstill
        \ElsIf{ your IL is at your LNP, with TYPE(IL)=Standstill and RIGHT(IL)=ON}
                 \State RIGHT $\pijl$ ON
                \State TYPE $\pijl$  Standstill
        \EndIf
   \end{algorithmic}
\end{algorithm}

\begin{figure}[t]
\vspace{.2cm} \centering
  \subfloat{\includegraphics[height=3cm]{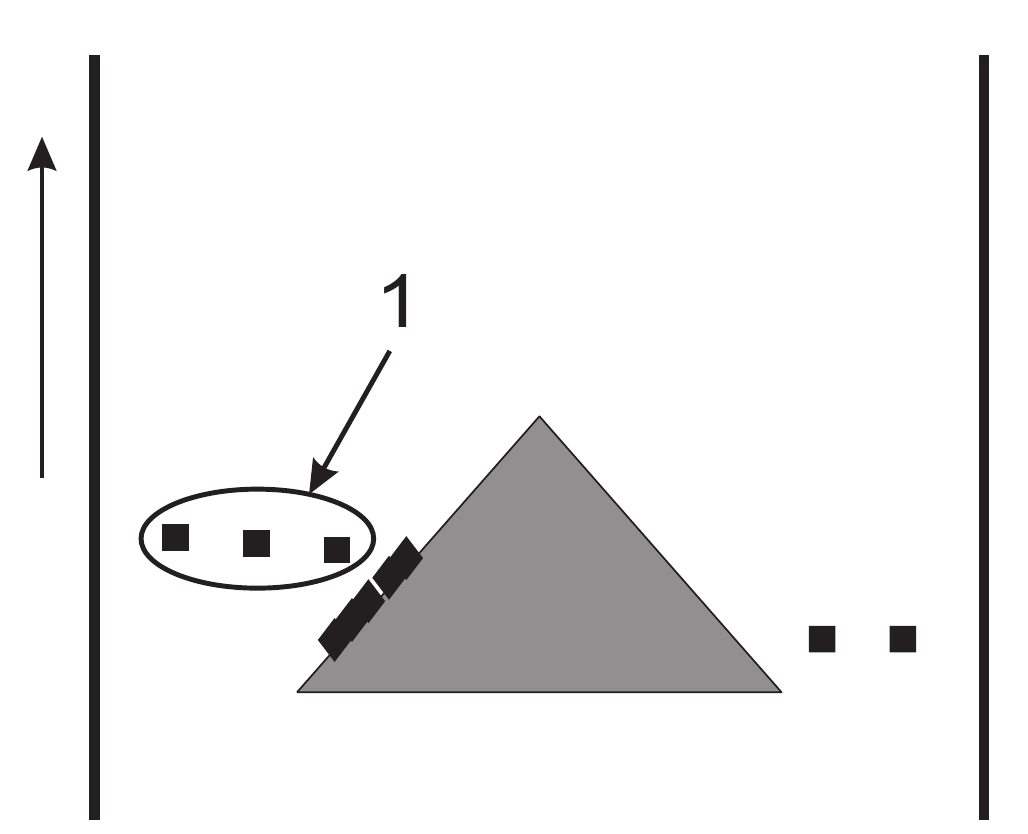}}\hfill
 \subfloat{\includegraphics[height=2.8cm]{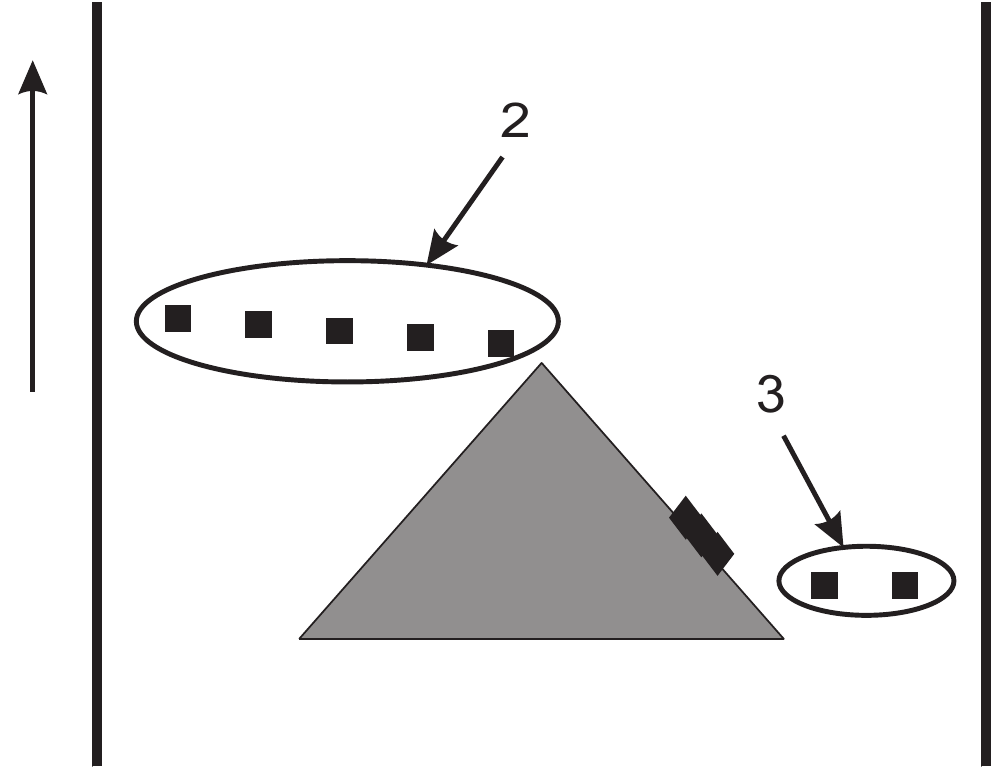}}
  \caption{Three typical situations where a subgroup of Standstill robots appears.
  Square = Standstill robot; diamond = Contour-Following robot. }\label{fig:redrob}
\end{figure}
\subsection{Switching from Run-Mode to Contour-Following}
The robot group is initialized with all robots in Run-Mode and all
parameters TOP, LEFT, and RIGHT turned OFF. The group advances
inside the strip in formation until it encounters an obstacle. In
the presence of obstacles, a robot switches {\bf from Run-Mode to
Contour-Following}
\begin{itemize}
\item if its Desired Position is located inside an obstacle and hence is
unreachable, or,
\item if its IL is Contour-Following and the forward path is obstructed by an
obstacle.
\end{itemize}

\subsection{Switching from Run-Mode to Standstill}
\label{sec:run2stand} If the above conditions are not satisfied, the
robot remains in Run-Mode and moves past the obstacle. Consider a
robot which passes on the left of an obstacle and has lost its IF at
its RNP because that robot started Contour-following around the
obstacle. This can only happen if the distance between the robot and
the obstacle decreased to a value less than $d$. While the robot
advances, its distance with the obstacle will eventually increase
again. The robot switches {\bf from Run-Mode to Standstill} if
Dist2O $> d$ and the corresponding obstacle is in the robot's fourth
quadrant. The robot's LEFT is turned ON. Similarly, if a Run-Mode
robot (apart from LSB) has no IL at its LNP and if Dist2O $> d$ and
the corresponding obstacle is in the robot's third quadrant, it is
forced to switch to Standstill. The robot's RIGHT is turned ON.

In practice, there exists a short delay between the moment the robot
senses that all conditions for standstill are satisfied and the
moment the robot effectively comes to a standstill. We take this
delay into account by introducing a small real constant $\epsilon$
such that each Run-Mode robot comes to a Standstill at a distance
from the obstacle with value in $(d,d+\epsilon)$.

To make sure subgroups of robots remain connected, we add two more
conditions to change from Run-Mode to Standstill:
\begin{itemize}
\item Switch to Standstill if your IF is at Standstill and its LEFT is ON.
\item Switch to Standstill if your IL is at Standstill and its RIGHT is ON.
\end{itemize}

\begin{figure}[t]
\centering
  \subfloat[]{\label{fig:a2}\includegraphics[height=2cm]{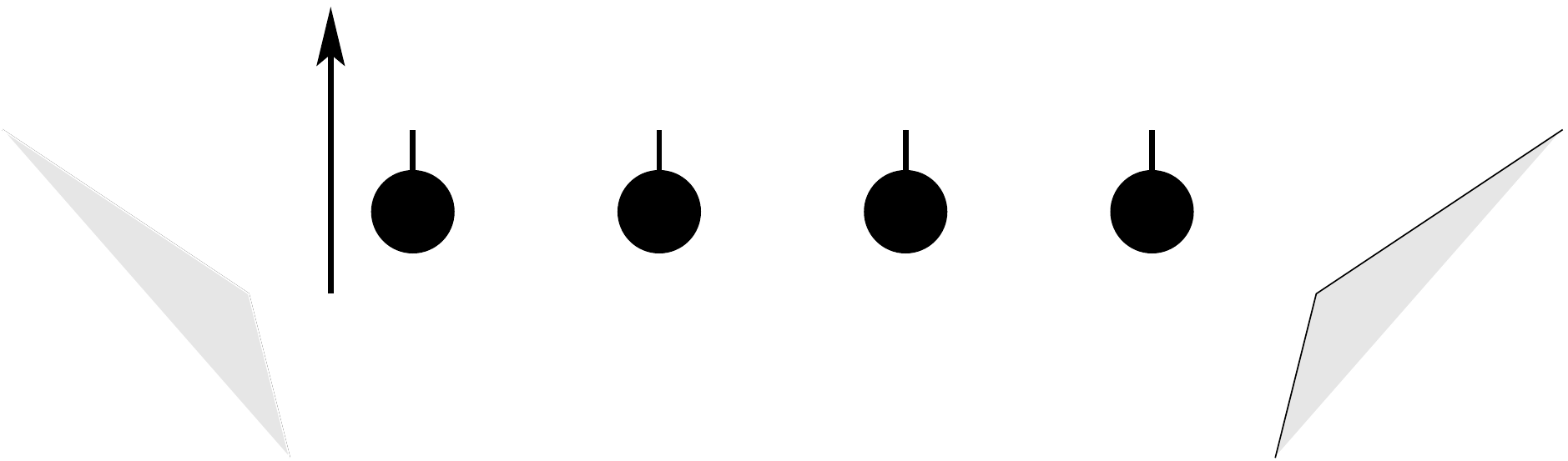}}
  \qquad \vspace{.3cm}
  \subfloat[]{\label{fig:b2}\includegraphics[height=2cm]{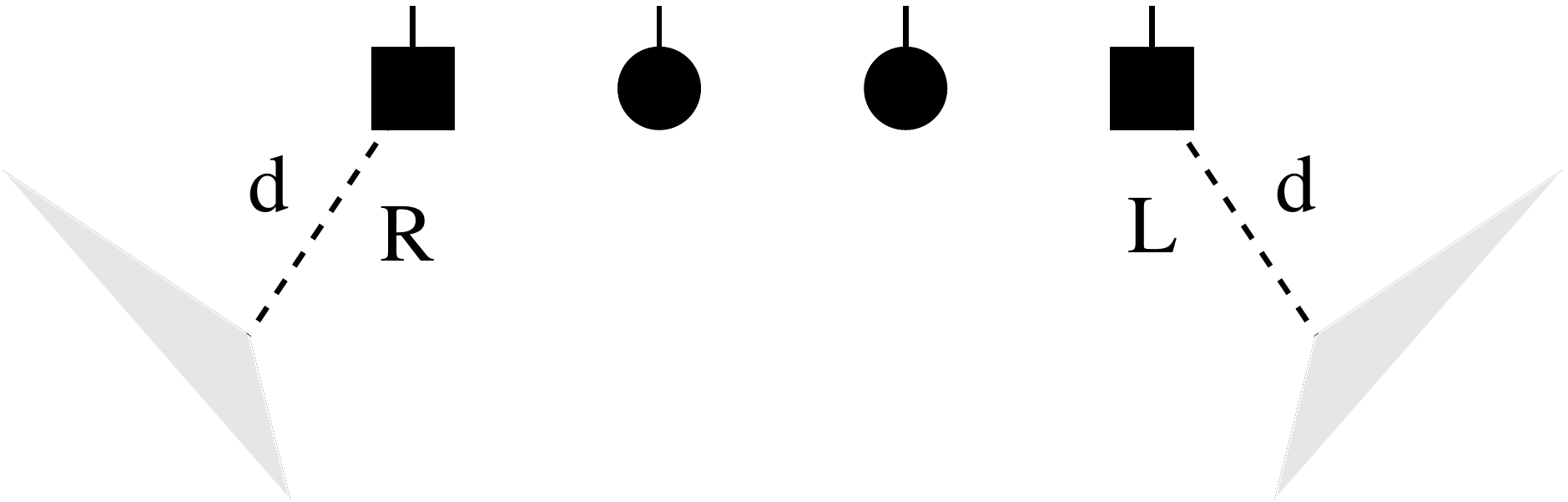}}  \qquad \vspace{.3cm}
\subfloat[]{\label{fig:c2}\includegraphics[height=2cm]{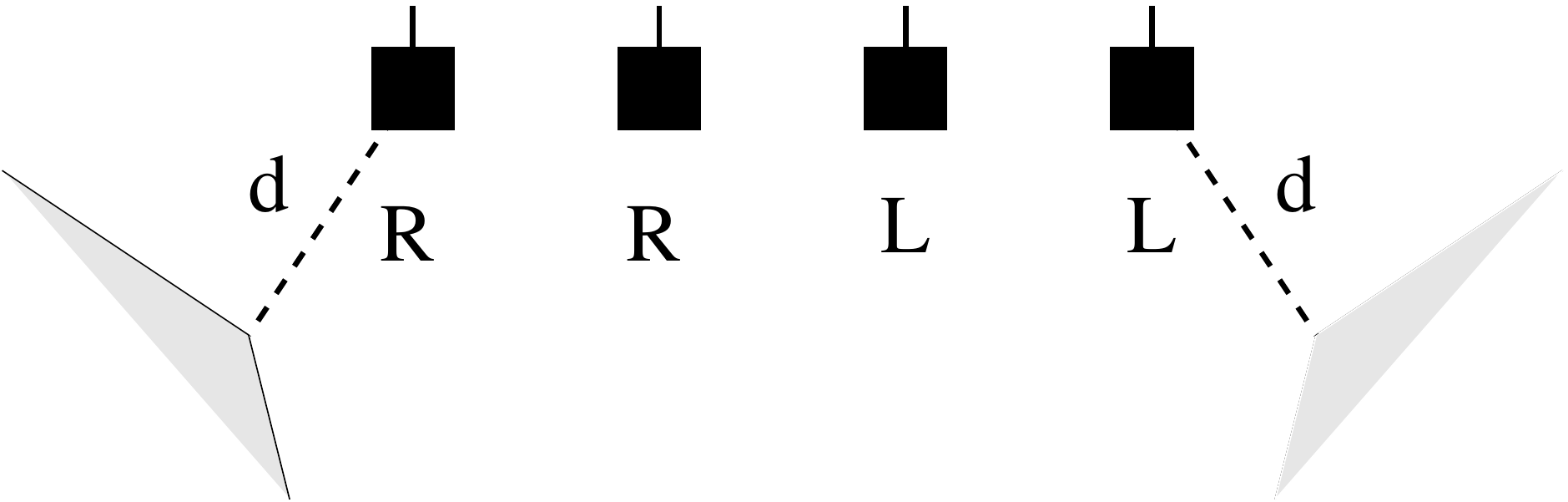}}
\qquad \vspace{.3cm}
\subfloat[]{\label{fig:d2}\includegraphics[height=2cm]{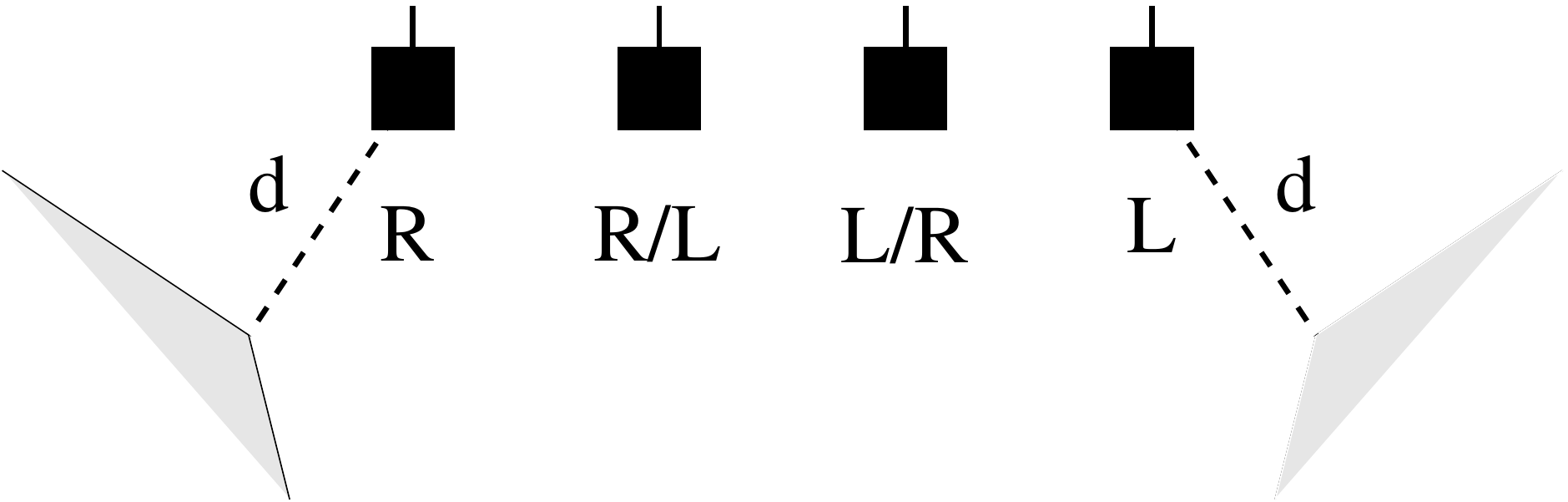}}
\qquad \vspace{.3cm}
\subfloat[]{\label{fig:e2}\includegraphics[height=2cm]{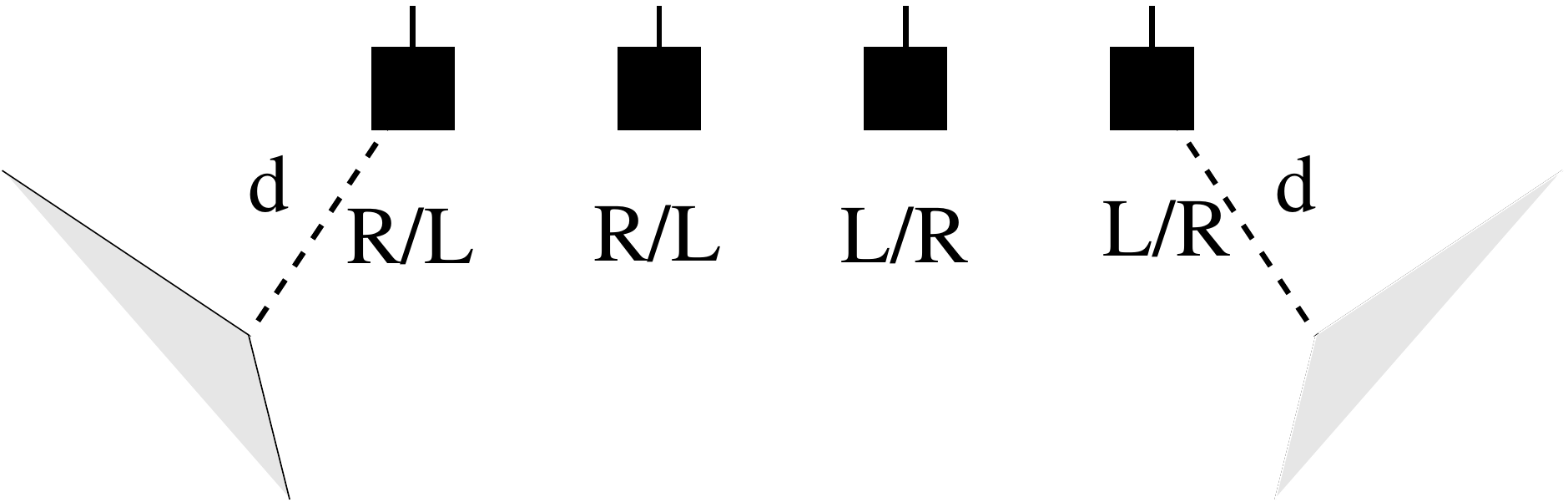}}
  \caption{A group of four Run-Mode robots moving between two obstacles and turning to Standstill
  with each robot activating both LEFT and RIGHT parameter.}\label{fig:nogmaken}
\end{figure}

In rare cases it is possible that a robot switches to standstill and
turns both the LEFT and RIGHT parameter ON. \figref{fig:nogmaken}
presents such a situation considering a robot subgroup of four
robots. The group is able to move through the gap created by two
obstacles (partially shown in the figure) (a). Because of the
symmetry of the configuration, the outer robots come to a standstill
at the same time. The leftmost/rightmost robot turns RIGHT/LEFT ON
(b). Shortly after, the two inner robots switch to Standstill . The
second robot turns RIGHT ON, the third activates LEFT (c). In
\algoref{alg:red} treating Standstill behavior it is shown that the
activation of a RIGHT or LEFT parameter propagates through an entire
group of Standstill robots (d). In the situation under consideration
this leads to all four robots having both LEFT and RIGHT parameters
turned ON (e).

\subsection{Switching from Contour-Following to Run-Mode or
Standstill}\label{sec:sit} The above rules lead to situations with
one or more robot groups at Standstill and other robots
Contour-Fol\-low\-ing around the obstacle. The key idea of the
algorithm is to make Contour-Following robots (re)join subgroups of
Standstill robots, in order to advance further inside the strip.
\figref{fig:redrob} shows the three possible situations for
subgroups of Standstill robots. (Robots indicated by a square are at
Standstill; robots depicted by a diamond are Contour-Following.) A
group of Standstill robots can be located

\begin{enumerate}
\item on the left of an obstacle (Situation $1$) \label{roodeen},
\item at a top of an obstacle (Situation $2$) \label{roodtwee},
\item on the right of an obstacle (Situation
$3$). \label{rooddrie}
\end{enumerate}
%in situations $1$ and $3$ until both groups are able to merge at the
%top of an obstacle. A Contour-Following robot encountering a subgroup
%in situation $2$ has to ignore this group.
A Contour-Following robot encounters a group of Standstill Robots by
detecting a Standstill robot with an available RNP or LNP. The
 robot then moves to this location and occupies it.

\subsubsection{Situation $3$}\label{sec:sitdrie}
If the position now occupied is the LNP of a Standstill robot, a
subgroup of Standstill robots in Situation~$3$ has been reached. In
general, each of these Standstill robots has its RIGHT parameter
turned ON. However, the Contour-Following robot first checks the
LEFT parameter of its new neighbor, in order to detect rare cases
like the situation discussed in \figref{fig:nogmaken}. If LEFT is
ON, this means that the group of Standstill robots under
consideration is not only located on the right side of an obstacle
(namely the obstacle the Contour-Following robot was moving around)
but also on the left side of another obstacle. The Contour-Following
robot takes on the Standstill mode and turns its LEFT parameter ON;
its RIGHT parameter remains OFF. This will cause a turning off of
the RIGHT parameter in the entire group of Standstill robots (see
\algoref{alg:red}), resulting in a Standstill group in Situation $1$
or $2$.

On the other hand, if the LEFT parameter of the new neighbor is OFF,
the Contour-Following robot simply turns to Run-Mode. The Standstill
robots respond by turning RIGHT off, resulting in a switch to
Run-Mode of the entire Standstill group (see \algoref{alg:red}).

Often the Contour-Following robot and the leftmost robot of the
Standstill group are not an original leader-follower pair. This
implies that they have to actively construct a new leader-follower
connection between each other so that they can proceed inside the
strip according to the Run-Mode part of the algorithm (see
\algoref{alg:Contour-Following}).
\begin{algorithm}
\caption{TYPE Contour-Following} \label{alg:Contour-Following}
    \begin{algorithmic}[5]
          \State Repeat
    \State {\it Move clockwise around obstacle},\\
     until
        \If{A robot of Contour-Following or Stand-Still type blocks your path}
        \State  Wait
        \ElsIf{Moving upward inside the strip $\land$ RNP(IL) is reachable $\land$ LEFT(IL) =
        ON $\land$ TOP(IL) = OFF}
        \State Move to RNP(IL)
            \If{RIGHT(IL)=ON}
                \State RIGHT $\pijl$ ON
                \State TYPE $\pijl$ Standstill
                %\State LEFT(IL) $\pijl$ OFF
            \Else
                   \State TYPE $\pijl$ Run-Mode
            \EndIf
            \ElsIf{Moving upward inside the strip $\land$ RNP(IL) is reachable $\land$ LEFT(IL) =
        ON $\land$ TOP(IL) = ON}
                 \State Keep on Contour-Following
        \ElsIf{Moving downward inside strip $\land$ you meet a robot with RIGHT=ON, TYPE Standstill and LNP reachable }
                \State Move to LNP of this robot
                \State Assign the new robot as your IF
                \State Assign yourself as the new robot's IL
                \If{LEFT(IF)=ON}
                    \State LEFT $\pijl$ ON
                    \State TYPE $\pijl$ Standstill
                   % \State RIGHT(IF) $\pijl$ OFF
                \Else
                    \State TYPE $\pijl$ Run-Mode
                \EndIf
            \ElsIf{ Moving downward inside strip $\land$ you meet a robot with TYPE Run-Mode}
            \State Stay at a fixed minimal distance from this robot
            \State and retrace your steps along obstacle
            if necessary
            \EndIf
   \end{algorithmic}
\end{algorithm}

\subsubsection{Situations $1$ and $2$}\label{sec:siteentwee}
In case the Contour-Following robot occupies the RNP of a Standstill
robot, the robot has to determine whether it reached a group in
Situation $1$ or Situation $2$. If it is concluded that a group in
Situation $1$ has been reached then a procedure similar to the one
for situation $3$ is performed. To obtain this procedure, just
switch ``LEFT'' and ``RIGHT'' in the description of
\secref{sec:sitdrie}. However, if Situation $2$ is detected, the
Contour-Following robot will abandon the RNP location to continue
around the obstacle in search of a Standstill group in Situation
$3$. The Standstill robot at the top of the obstacle will turn its
TOP parameter ON, ensuring other Contour-Following robots
encountering the robot pass it without checking the Standstill
group.

The distinction between Situation $1$ and $2$ is easy to detect. The
Contour-Following robot, located at the RNP of its IL, checks the
direction along which it detects the obstacle it was moving around.
If this direction belongs to the fourth quadrant of the
Contour-Following robot's surroundings, Situation $1$ holds; if it
belongs to the third quadrant, Situation $2$ is detected.
\begin{figure}[t]
\centering
  \includegraphics[height=3cm]{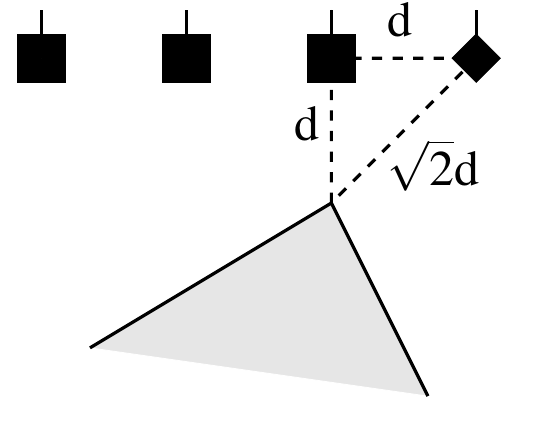}
  \caption{A Contour-Following robot (diamond) has reached the RNP
  of a Standstill robot (square) at the top of an obstacle.
  The figure depicts the largest possible distance between the Contour-Following
  robot and the obstacle.}\label{fig:atthetop}
\end{figure}

Remark that we do not allow the robot to record data on the
environment. A robot is not able to remember which obstacle it was
moving around a time instant earlier. However, if a
Contour-Following robot joins a standstill group in situation $1$ or
$2$ this information is required, as explained above. Therefore we
introduce the assumption which leads to a minimum inter-obstacle
distance.

\begin{ass}
\label{ass:distance} When a Contour-Following robot joins a
Standstill group, the obstacle the robot was moving around is the
nearest obstacle of all obstacles detected by the robot, i.e. the
obstacle belonging to the robot's measured Dist2O.
\end{ass}

%that a robot only keeps track of the {\it shortest} distance between
%itself and {\it all} obstacles surrounding it (Dist2O).
%Consequently, in order to ensure that the Contour-Following robot
%detects the correct direction, the obstacle it has been moving
%around needs to be the nearest obstacle when the robot is .
The largest possible distance between a robot that just joined a
standstill group and the obstacle it was moving around is depicted
in \figref{fig:atthetop}. Its size is computed to be $\sqrt{2}d$. It
follows from \assref{ass:distance} that all other obstacles are
located further away from the robot. This imposes a lower bound on
the inter-obstacle distances: the minimum allowable interdistance
equals $2 \sqrt{2}d$.

\subsection{Switching from Standstill to Run-Mode}
The previous sections showed that whenever a robot switches to
Standstill, it also turns ON one of the parameters LEFT and RIGHT.
These parameters determine the behavior of the Contour-Following
robot joining a Standstill robot.

A Standstill robot will switch back to Run-Mode if both its LEFT and
RIGHT parameters are turned OFF. Each Standstill robot copies the
behavior of its left neighbor to turn the RIGHT parameter on or off,
and similarly copies the behavior of its right neighbor to turn LEFT
on or off.

\begin{algorithm}
\caption{TYPE Standstill} \label{alg:red}
    \begin{algorithmic}[5]
        \State Repeat
       \State {\it Stand still and}
            %\If{your IF is at RNP but your IL is not at LNP $\land$ LEFT(IF) = OFF}
%                \State RIGHT $\pijl$ OFF
%            \ElsIf{your IL is at LNP but your IF is not at RNP $\land$ RIGHT(IL) = OFF}
%                \State LEFT $\pijl$ OFF
            \If{your IL is at your LNP $\land$ RIGHT(IL) = ON/OFF}
                \State RIGHT $\pijl$ ON/OFF
            \ElsIf{your IF is at your RNP $\land$ LEFT(IF) =ON/OFF}
                \State LEFT $\pijl$ ON/OFF
            \EndIf\\
        until
        \If{LEFT = OFF $\land$ RIGHT = OFF}
            \State TYPE $\pijl$ Run-Mode
        \ElsIf{ LEFT= ON $\land$ your IL is at your LNP $\land$ a new robot in Run-Mode appears at RNP}
            \State Assign the new robot as your IF
            \State Assign yourself as the new robot's IL
            \State \Comment (comment) This reconnects two subgroups at
                the top of an obstacle
        \EndIf
   \end{algorithmic}
\end{algorithm}

\begin{figure*}[ht]
\centering
  \subfloat[]{\label{fig:a1}\includegraphics[height=2.7cm]{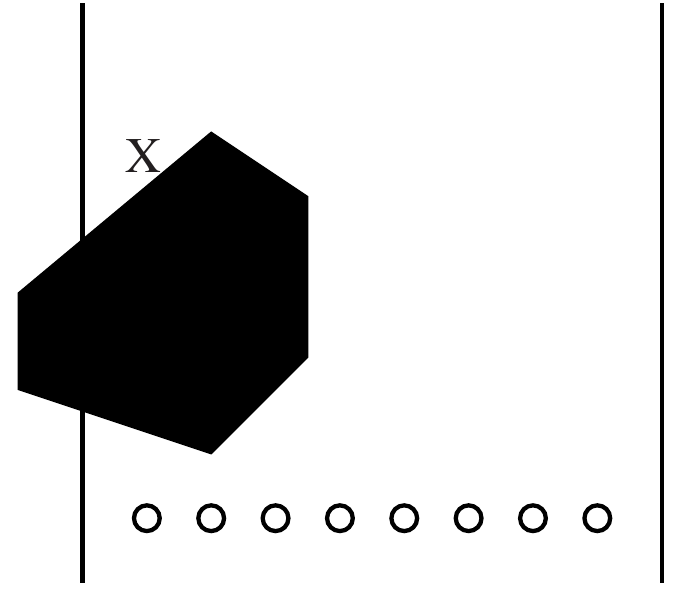}}
  \qquad \quad
  \subfloat[]{\label{fig:b1}\includegraphics[height=2.7cm]{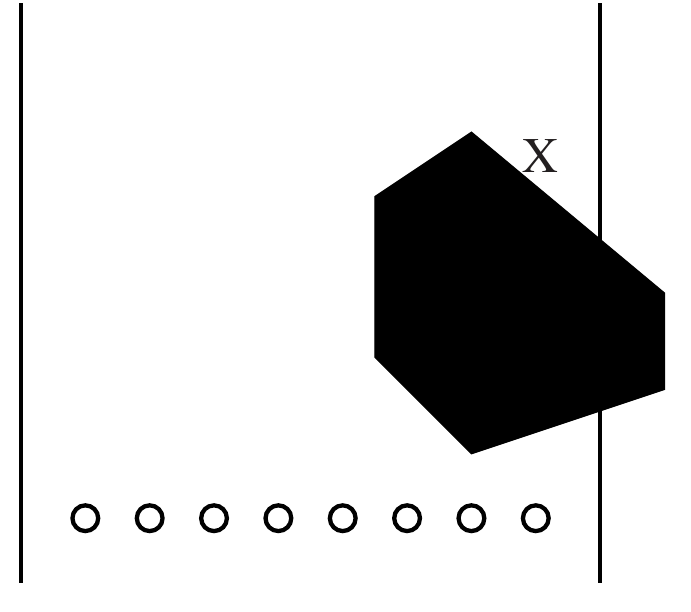}}
  \qquad \quad
\subfloat[]{\label{fig:c1}\includegraphics[height=2.7cm]{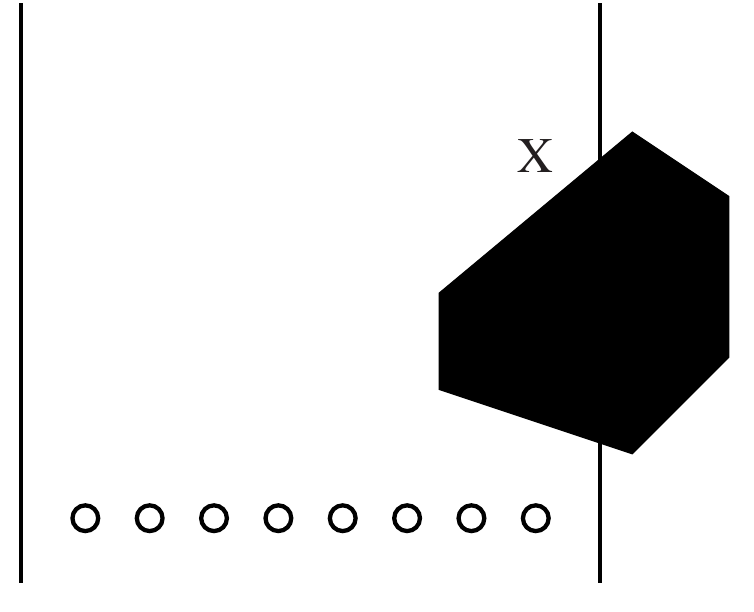}}
\qquad \quad
\subfloat[]{\label{fig:d1}\includegraphics[height=2.7cm]{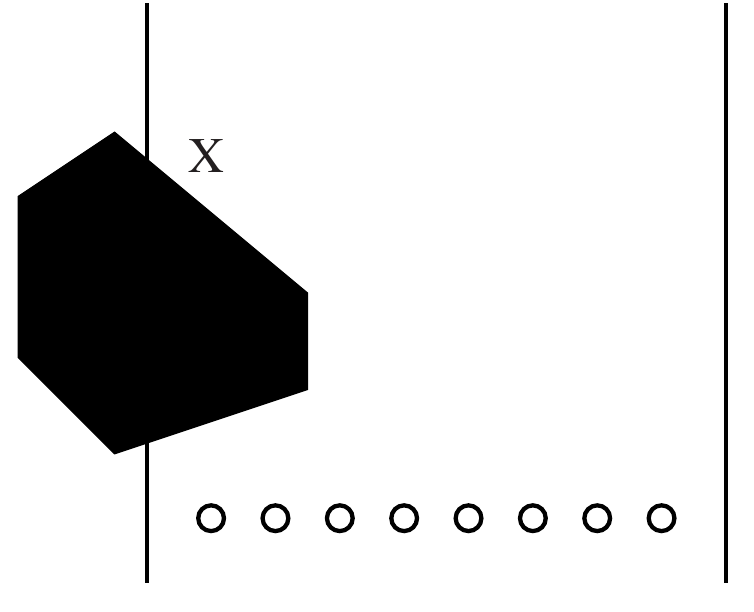}}
  \caption{Four distinct cases of an obstacle located on the strip boundary.
  The two situations on the right demand an extension of the algorithm,
  the other two do not.}\label{fig:obstonbound}
\end{figure*}

\subsection{Obstacles on the strip boundary}\label{sec:obstbound}
As mentioned in the introduction, obstacles are allowed to be
located on a strip boundary. The algorithm will treat such obstacles
by sending (part of) the robot group around the obstacle {\it
outside of the strip}. %If the strip boundary is solid, the robots
%are unable to cross it, which may lead to a dead-lock situation in
%the case of obstacles located on the strip boundary. Therefore, we
%exclude obstacles which are touching a solid strip boundary.
To execute this maneuver, the algorithm of the Interior Robots does
not need to be changed. Both LSB and RSB are given an algorithm
which slightly differs from the other robots. When its forward path
is obstructed by an obstacle, the LSB (RSB) robot moves clockwise
(counterclockwise) around the obstacle {\it outside} of the strip,
until it reaches its original horizontal position. This location is
called the {\it reentry point} and is indicated by a letter X in
\figref{fig:obstonbound}. In order to measure this position the
robot is equipped with GPS. Once the reentry point has been reached,
one of four possible configurations is detected, as depicted in
\figref{fig:obstonbound}, each requiring a different treatment.

First, consider \figref{fig:a1}. The obstacle will cause the robot
group to split between the third and the fourth robot. The LSB moves
clockwise around the obstacle and reaches its reentry point.
Resuming its basic algorithm, it moves parallel to the strip
boundary until the distance between the obstacle and itself exceeds
the preset threshold~$d$. This creates a situation similar to
situations $1$ or $2$ of \figref{fig:redrob}. Interior Robots $2$
and~$3$, moving clockwise around the obstacle, will reconnect to the
LSB or, if the LSB is located at the top of the obstacle, will
continue along the obstacle to expand the robot subgroup located on
the right side of the obstacle. Without further adjustments the
basic algorithm is able to tackle this situation as required.

In this respect, the configuration in \figref{fig:b1} causes similar
behavior. All robots move along the left of the obstacle, except for
the RSB. The RSB reaches its reentry point after which it moves
forward up to the preset distance $d$ with the obstacle. The
situation attained is similar to situation $3$ of
\figref{fig:redrob}. Again, no further adjustments are needed to let
the basic scanning algorithm finish the job satisfactorily.

The configurations depicted in \figref{fig:c1} and \ref{fig:d1} do
{\it not} satisfy the pattern observed in the previous two cases.
When the RSB reaches the reentry point, a previously unencountered
situation is created. The slope of the obstacle's boundary near the
reentry point does not allow any of the situations $1$, $2$, or $3$.
The algorithm of the RSB needs to be extended. After reaching its
reentry point, we let the RSB sense the orientation of the slope of
the obstacle boundary. The robot does this by checking the direction
along which it detects the obstacle under consideration. If this
direction belongs to the fourth quadrant of the RSB, as depicted in
\figref{fig:c1}, the robot abandons its standard behavior and
remains positioned at the reentry point instead. It waits for its IL
to appear within its sensor range. This IL will end up as the
rightmost robot of a Standstill robot group in situation $1$. The
RSB is able to detect this and can reoccupy the RNP of its IL. The
RSB completes the robot group and switches back to its standard
behavior.

Finally, the situation in (\figref{fig:d1}) is addressed. Similar to
the previous case, after sensing the obstacle boundary slope at the
reentry point, the LSB switches to an extension of its original
program. It moves forward until the distance between the obstacle
and itself exceeds the preset threshold, at which point it waits
until all robots which followed it clockwise around the obstacle
have passed. These robots expand the robot group on the right side
of the obstacle. The LSB retraces it steps towards the obstacle
until it reaches the LNP of the leftmost robot of the (right) robot
group. The LSB connects to this robot completing the robot group.
The obstacle has been tackled successfully and the sweeping of the
strip can continue.

\subsection{Simulation results}\label{sec:sim}
The algorithm presented in this section has been successfully
implemented in Matlab code. Our program treats diverse obstacle
configurations. \figref{fig:obstsim} is the result of one such run
of the program where a situation with two obstacles is considered.
The figure displays $8$ snapshots taken during the simulation, with
reading order left to right, top to bottom. To better verify the
capabilities of the program, the output of some runs of the program
have been recorded in the form of videos. These videos can be
retrieved at http://www.systems.ugent.be/videos/. In these animated
simulations a color code has been used to indicate the status of
each robot, instead of different shapes of the symbols representing
the robots.

The simulations allow us to quantify the time gain obtained from
using a robot team compared to single-robot coverage. The
single-robot coverage strategy we take into consideration for
comparison is the algorithm executed by the LSB robot in the present
paper. In a single-robot setting, this robot sweeps a strip of one
robot wide using GPS, and whenever it encounters an obstacle, it
moves clockwise around it until it reaches its assigned strip again.

We conducted simulations for both a group of $10$ robots and a
single robot. In case of an empty strip the $10$-robot group sweeps
the strip $10$ times faster than the single robot, as expected. In
the presence of obstacles the time gain is reduced: our simulations
show that the group of $10$ robots finishes a strip $6-8$ times
faster than the single robot. The exact value of the time gain
depends on the obstacle configuration under consideration.

\section{Proof of coverage}\label{sec:proof}
In \secref{sec:strip} it was shown how the surface $A$ of a strip
was covered when the robot team tracks a set $V$ of parallel lines
with interdistance $d$. In the presence of obstacles, this
proposition holds with $V$ and $A$ replaced by $V \setminus
\mathcal{P}$ and $A \setminus \mathcal{P}$ respectively. We add the
following definition.
\begin{defi}
A set consisting of all points with the same $x$-value belonging to
$V \setminus \mathcal{P}$ is called a \emph{ basic robot track}.
Each connected subset of a basic robot track is called a \emph{
section} of a basic robot track.
\end{defi}
We conclude that, in order to prove coverage of the entire area, it
is sufficient to prove coverage of all basic robot track sections.
This is the subject of the present section.  For simplicity we
restrict to situations where
\begin{itemize}
    \item the basic robot tracks of the LSB and RSB are not obstructed by
obstacles,
    \item all obstacles are convex with exactly one top.
\end{itemize}
%We start by proving $2$
%lemma's.

\begin{lem}\label{theor:green}
Each robot in Run-Mode moves along a trajectory with constant
x-coordinate.
\end{lem}
\begin{pf}
The robot group under consideration is finite, which implies the
existence of one or more disjoint finite subgroups of Run-Mode
robots in a straight-line formation and with consecutive indices. If
a robot is the leftmost robot of a Run-Mode subgroup, it satisfies
at least one of the following conditions: the robot
\begin{itemize}
    \item has no IL,
    \item does not sense its IL, or
    \item senses its IL, which is Contour-Following.
\end{itemize}
This is proven by contrapositive: assume that for a robot none of
the above three conditions is satisfied. Then one of the following
situations holds:
\begin{enumerate}
    \item The robot senses its IL which is at Standstill.
    Consequently the robot turns to Standstill, and hence is not in Run-Mode.
    \item The robot senses its IL which is in Run-Mode.
    Consequently the robot is also in Run-Mode but is not the leftmost robot of a Run-mode subgroup.
    \end{enumerate}
From the three conditions above and the definition of Run-Mode (see
\secref{sec:robpar}), it follows that the leftmost robot of a
Run-Mode subgroup moves with constant x-coordinate. The remaining
robots of the Run-Mode subgroup copy this movement, since they all
sense their respective IL in Run-Mode.
% it moves forward
%with constant x-coordinate. Otherwise, for each robot pair
%$(i,\text{IF}(i))$ that is in Run-Mode ($i=1,\ldots,N-1,$), robot
%IF($i$) occupies the RNP of robot $i$. In other words, if robot $i$
%follows the trajectory $t \mapsto (x(t),y(t))$ in Run-Mode, then
%IF($i$) traces the trajectory $t \mapsto (x(t)+d,y(t))$.
\end{pf}

\begin{lem}\label{theor:yellow}
Consider a scanning strip with boundaries at $x=x_L$ and $x=x_R$,
with $x_L < x_R$. Let $P$ be an obstacle with $x_{min}$ the
x-coordinate of its leftmost point(s) and $x_{max}$ the x-coordinate
of its rightmost point(s). Eventually all robots moving along a
basic robot track section with the x-coordinate belonging to
$[\max(x_L,x_{min}), \min(x_R,x_{max})]$ and the end-point at $P$,
will switch to Contour-Following around $P$.
\end{lem}
\begin{pf}
The only situation where a robot with the properties described above
will {\it not} start Contour-Following around $P$ is when it
switches from Run-Mode to Standstill before reaching the obstacle.
We investigate if this situation can occur. Standstill can only be
caused by the presence of an obstacle $Q_i$ (different from $P$
because of the convexity assumption). The Stand-till robot belongs
to a subgroup in one of the situations $1$, $2$ or $3$ presented in
\secref{sec:sit}, related to an obstacle $Q_i$ characterized by
leftmost and rightmost points $(x_{i,min}$ and$x_{i,max})$. The
robot remains at Stand-still if the corresponding Standstill group
does not get joined by a robot Contour-Following around $Q_i$. It is
easy to see that this implies that not all robots with coordinate
 $x \in [\max(x_L,x_{i,min}), \min(x_R,x_{i,max})]$ start
Contour-Following around $Q_i$ (or that at least one
Contour-following robot gets obstructed by a persistent Standstill
group belonging to an obstacle $Q_j \neq Q_i$).

% takes on
%Standstill behavior before or during Contour-Following around $Q_i$
%and do not switch back. (Robots that are Contour-Following around
%$Q_i$ and join a Standstill group different from the group under
%consideration are allowed.) It can be easily proved that this
%requires the existence of a robot group at Standstill related to an
%obstacle different from $Q_i$ and $P$.

This observation shows that the original assumption at obstacle $P$
(i.e. not all robots with x-coordinate in $[\max(x_L,x_{min}),
\min(x_R,x_{max})]$ will start contour-Following around $P$) leads
to the same situation at an obstacle $Q_i$ different from $P$.
Consequently, the initial assumption implies the existence of an
infinite series of different obstacles. Since the number of
obstacles is assumed to be finite, the original assumption is not
valid. This concludes the proof.
%
%
%
%This implies that each robot with $x \in [\max(x_L,x_{i,min}),
%\min(x_R,x_{i,max})]$ belongs to a subgroup in situation $1$, $2$ or
%$3$ related to an obstacle different from $P$ and $Q_i$. To
%guarantee the persistent existence of the Standstill robots, the
%above reasoning needs to be repeated an infinite number of times.
%This requires an infinite number of obstacles.  our assumption is
%not valid. It follows that all robots with $x \in
%[\max(x_L,x_{min}), \min(x_R,x_{max})]$ do not remain at Standstill
%but eventually will move forward with constant x-coordinate until
%they reach the obstacle $P$, where they start to perform
%Contour-Following around $P$.
 \end{pf}
We now present the proof of coverage.
\begin{thm}\label{theor:dead}
If the minimum inter-obstacle distance is $2 \sqrt{2}d$, then
%\begin{enumerate}
%    \item the robot group does not encounter dead-lock situations;
%    \item
the algorithm covers all basic robot tracks completely.
%\end{enumerate}
\end{thm}
\begin{pf}
At the initialization of the algorithm, the robot group has assumed
the formation defined in \secref{sec:robfor}. Each robot is located
at the start of a basic robot track. It follows from
\lemref{theor:green}, that at the beginning of the algorithm each
robot moves in Run-Mode along its basic robot track. A robot
abandons the basic robot track it was moving on, if and only if it
has switched to Contour-Following. We have to prove that
\begin{itemize}
    \item every Contour-Following robot switches back to Run-Mode if and only if
it is located on a basic robot track,
    \item each section of a basic robot track is traced once,
\end{itemize}

Consider an obstacle $P$ characterized by $x_{min}$ and $x_{max}$
The robots moving along basic robot tracks with $x \in [x_{min},
x_{max}]$ will reach $P$ and switch to Contour-Following (cfr.
\lemref{theor:yellow}). The two robots located on the basic robot
tracks with $x \in [x_{min}-d,x_{min})$ or $x \in (x_{max},
x_{max}+d]$ move past the obstacle. Call these robots $R_L$ and
$R_R$ respectively. (These robots exist, due to the assumption of
unobstructed robot tracks for LSB and RSB.) These two robots come to
a standstill, when the distance between itself and obstacle $P$
satisfies the following conditions (cfr. \secref{sec:run2stand}):
\begin{itemize}
    \item has a value in $(d,d+\epsilon), \: \epsilon \ll 1$ ,
    \item it is the shortest of all distances between the robot and all obstacles surrounding it.
\end{itemize}
Satisfaction of the second condition is guaranteed by the theorem's
assumption on the minimum inter-obstacle distance. At standstill,
both robots $R_L$ and $R_R$ lack a neighboring robot on the side
where the obstacle is located. Since the robots are located more
than $d$ distance units away from any obstacle they have either an
unoccupied LNP (in the case of $R_R$) or unoccupied RNP (for $R_L$)
that is not obstructed by an obstacle. The robots $R_L$ and $R_R$
are located on basic robot tracks, so their LNP and RNP are located
on basic robot tracks as well. Since $P$ has exactly one top, one of
the following configurations has been attained:
\begin{enumerate}
    \item LNP of $R_R$ at the top, RNP of $R_L$ belonging to situation $3$ (as defined in \secref{sec:sit});
    \item LNP of $R_R$ belonging to situation $1$, RNP of $R_L$ at the top;
    \item LNP of $R_R$ belonging to situation $1$, RNP of $R_L$ belonging to situation $3$;
\end{enumerate}
In each of the three cases, a Contour-Following robot has the
opportunity to occupy either a free LNP or a free RNP. From
\lemref{theor:yellow} and the description of the algorithm in
\secref{sec:alg} it follows that the number of Contour-Following
robots available is necessary and sufficient to occupy the free
RNPs/LNPs. The theorem's assumption ensures that the
Contour-Following robots are able to tackle the top of the obstacle
which guarantees that every available LNP will be reached by a
Contour-Following robot, as explained in \secref{sec:sit}. When a
Contour-Following robot has obtained the free RNP/LNP, its sensors
are covering the part of its basic robot track that lies between
itself and the obstacle it was moving around. The robot turns to
Run-Mode and starts tracing the rest of the respective basic robot
track section. The entire reasoning followed for $R_L$ or $R_R$ can
now be repeated for the Contour-Following robot which has switched
back to Run-Mode. Every time a Contour-Following robot comes to
Standstill it creates an available RNP or LNP to be occupied by a
next Contour-Following robot, until the two robot groups merge at
the top of the obstacle. It follows that each section of a basic
robot track is traced by precisely one robot.

% Each Standstill robot
%will later switch back to Run-Mode and continue tracing its basic
%robot track. Each Contour-Following robot will occupy the LNP or RNP of
%a Standstill robot in Situation $1$ or $2$.
%
% As shown in \lemref{theor:dead} the Wall
%Following robot will switch to Run-Mode after some time and trace
%its newly assigned basic robot track. The theorem also showed that
%each Contour-Following robot will be assigned to one robot track
\end{pf}

\section{Conclusion}
This paper describes an algorithm for multi-robot coverage in an
unknown environment. The algorithm uses robot formations, which
\begin{itemize}
    \item reduces the need for extensive sensor capabilities,
    \item keeps radio communication between robots at a minimum,
    \item enables us to treat both coverage and pursuit-evasion
    missions.
\end{itemize}
The robots scan the environment along predefined strips. Information
on location of the strips is loaded into the memory of both outer
robots LSB and RSB at the beginning of the algorithm and remains
unchanged. Hence, a cellular decomposition of the environment that
is dynamically created during the algorithm, is not called for. Such
a decomposition requires more memory capacity of the robots and more
inter-robot communication. Dividing the area into fixed strips
requires an intelligent algorithm to pass the obstacles located in
the environment. Our algorithm successfully treats all possible
configurations of convex obstacles. As demonstrated in simulations,
the robot team is able to pass multiple obstacles simultaneously as
well as obstacles located on the strip boundary. A proof of coverage
concludes the paper.

\section*{Acknowledgment}
This paper presents research results of the Belgian Network DYSCO
(Dynamical Systems, Control, and Optimization), funded by the
Interuniversity Attraction Poles Programme, initiated by the Belgian
State, Science Policy Office. The scientific responsibility rests
with its authors.

\bibliographystyle{model1-num-names}
\bibliography{robotbib}

\end{document}